\documentclass[sn-basic]{sn-jnl}
\pdfoutput=1

\usepackage{etoolbox}
\makeatletter
\patchcmd{\ps@headings}
{\hbox to \hsize{\hfill Springer Nature 2021 \LaTeX\ template\hfill}}
{\hbox to \hsize{\hfill}}
{}
{}
\patchcmd{\ps@titlepage}
{\hbox to \hsize{\hfill Springer Nature 2021 \LaTeX\ template\hfill}}
{\hbox to \hsize{\hfill}}
{}
{}
\patchcmd{\ps@headings}
{\hbox to \hsize{\hfill Springer Nature 2021 \LaTeX\ template\hfill}}
{\hbox to \hsize{\hfill}}
{}
{}
\makeatother

\usepackage[colorinlistoftodos]{todonotes}
\setuptodonotes{inline}

\usepackage[font=small]{caption}

\newcommand{\totalN}{212}
\newcommand{\cccN}{25}
\newcommand{\ccdN}{29}
\newcommand{\ccN}{25}
\newcommand{\cdN}{26}
\newcommand{\dcN}{25}
\newcommand{\ddcN}{26}
\newcommand{\dddN}{27}
\newcommand{\ddN}{29}
\newcommand{\filteredN}{181}
\newcommand{\fcccN}{23}
\newcommand{\fccdN}{22}
\newcommand{\fccN}{20}
\newcommand{\fcdN}{24}
\newcommand{\fdcN}{22}
\newcommand{\fddcN}{25}
\newcommand{\fdddN}{23}
\newcommand{\fddN}{22}
\newcommand{\completionTime}{7.36 minutes}
\newcommand{\totalComp}{2.32}
\newcommand{\filterToTotalPercent}{85.38\%}

\newcommand{\twoTotalN}{250}
\newcommand{\twoCcN}{46}
\newcommand{\twoCccN}{40}
\newcommand{\twoDcN}{42}
\newcommand{\twoNccN}{41}
\newcommand{\twoNcccN}{41}
\newcommand{\twoNdcN}{40}
\newcommand{\twoCompletionTime}{12.38 minutes}
\newcommand{\twoTotalComp}{1.99}

\jyear{2021}%

\theoremstyle{thmstyleone}%

\theoremstyle{thmstyletwo}%

\theoremstyle{thmstylethree}%

\begin{document}

\title[Actively learning to learn causal relationships]{Actively learning to learn causal relationships}

\author*[1]{\fnm{Chentian} \sur{Jiang}}\email{chentian.jiang@ed.ac.uk}
\author[1]{\fnm{Christopher G.} \sur{Lucas}}\email{clucas2@inf.ed.ac.uk}
\affil*[1]{\orgdiv{School of Informatics}, \orgname{University of Edinburgh}, \orgaddress{\street{10 Crichton Street}, \city{Edinburgh}, \postcode{EH8 9AB}, \country{United Kingdom}}}

\abstract{
  How do people actively learn to learn? That is, how and when do people choose actions that facilitate long-term learning and choosing future actions that are more informative? We explore these questions in the domain of active causal learning.
  We propose a hierarchical Bayesian model that goes beyond past models by predicting that people pursue information not only about the causal relationship at hand but also about causal overhypotheses---abstract beliefs about causal relationships that span multiple situations and constrain how we learn the specifics in each situation.
  In two active ``blicket detector'' experiments with 14 between-subjects manipulations, our model was supported by both qualitative trends in participant behavior and an individual-differences-based model comparison. 
  Our results suggest when there are abstract similarities across active causal learning problems, people readily learn and transfer overhypotheses about these similarities. Moreover, people exploit these overhypotheses to facilitate long-term active learning.
  }

\keywords{causal learning, active learning, transfer learning, overhypotheses}

\maketitle

A key feature of human cognition is that when we learn, we often acquire knowledge and skills we can use in the future, improving our performance and future learning \cite[e.g.,][]{gickAnalogicalProblemSolving1980,schulzCausalLearningDomains2004}.
For instance, cooking one dish can help us learn how to work with ingredients in another dish, playing with the user interface of one smartphone can also help us learn to navigate another smartphone, and practicing one musical instrument can help us learn a new instrument more quickly. In each case, we are able to learn general patterns and principles, such as common chord progressions and conventions in phone interfaces, that go beyond a particular context or problem. This enables us to focus on the novel aspects of future problems and thereby learn more efficiently.
This ``learning to learn'' depends on \textit{overhypotheses} \citep{goodmanFactFictionForecast1955,kempLearningOverhypothesesHierarchical2007}---abstract beliefs that span multiple situations and constrain how we learn the specifics in each situation.

However, it is not well understood how and when we seek out the evidence needed to learn overhypotheses. Do people tend to focus narrowly on learning the task at hand, so that learning overhypotheses happens incidentally? Or do we preferentially choose actions to update our overhypotheses about the abstract nature of families of systems and problems? When we update our overhypotheses in light of new evidence, does that in turn facilitate more informative actions in a new situation? These are questions about how we \textit{actively} learn to learn. We explore them in the domain of active causal learning.

Active learning is useful because it allows people to take control of their own learning and seek information that is most helpful given their beliefs and uncertainty \citep{gureckisSelfDirectedLearningCognitive2012}. In causal learning, actions or \textit{interventions} can further provide information that is unavailable under observation alone; this information is critical for disambiguating causal relationships \citep{pearlCausality2009}.
Both of these benefits of interventions have been shown for adults and children, and have been neatly formalized in computational models of active causal learning \citep[e.g.,][]{steyversInferringCausalNetworks2003,bramleyConservativeForgetfulScholars2015,coenenStrategiesInterveneCausal2015,cookWhereScienceStarts2011}.

\begin{figure}[t]
  \centering
  \includegraphics[width=0.6\textwidth]{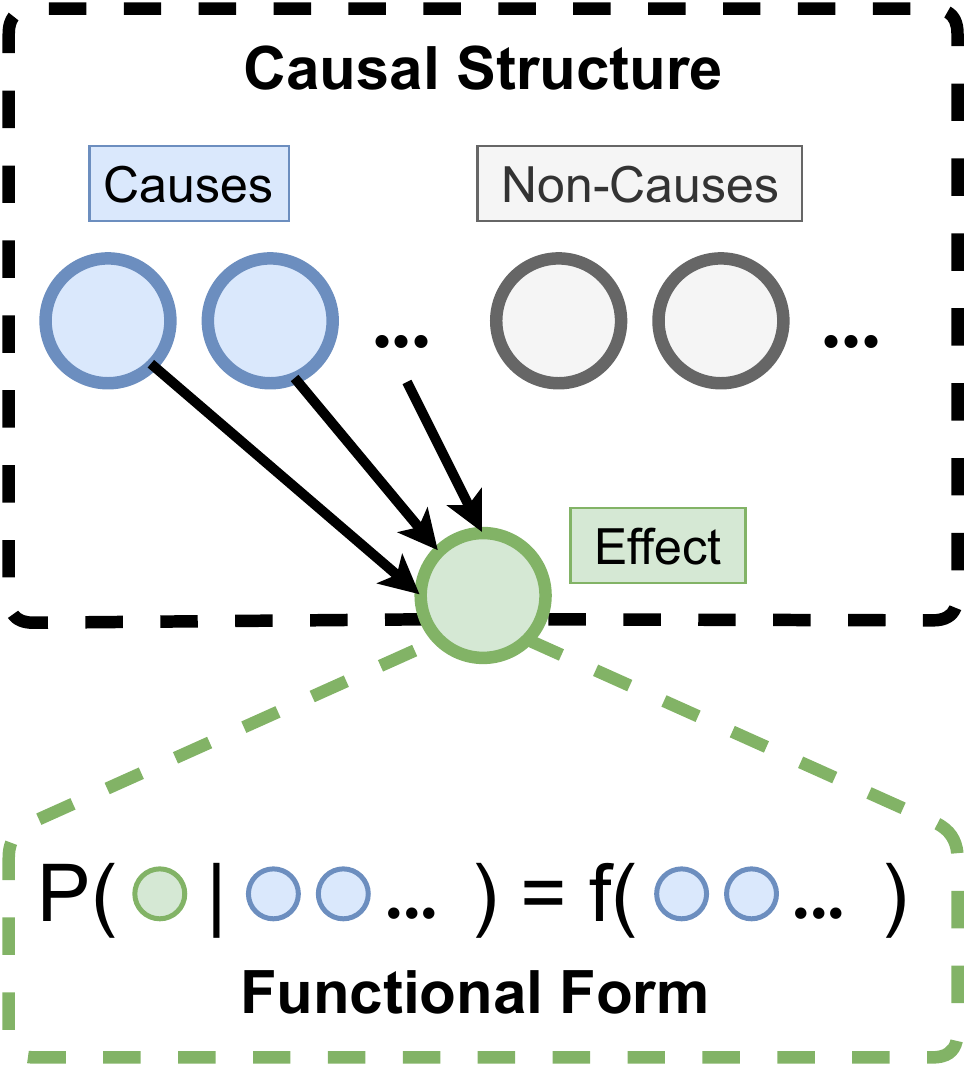}
  \caption{Causal graph. The \textit{causal structure} defines what variables are causes and effects of other variables. The \textit{functional form} is a function of its causes and it defines the conditional probability of the effect given its causes.}
  \label{fig:causal_graph}
\end{figure}

These past models have focused on how people learn about \textit{causal structure}, which defines what variables are causes and effects of other variables  %
(Fig.~\ref{fig:causal_graph}).
Consider an example where a child conducts a small science experiment to test which batteries in their drawer are good or bad. The child devises an intervention strategy based on inserting batteries into a simple circuit with an LED light. The LED is known to illuminate when at least one good battery is in the circuit. Here the child has set up a causal system where the variables are the batteries' presence in the circuit and the LED's illumination. Common sense about electrical systems dictates that the LED illumination is the only candidate for being an effect. Therefore, the child is trying to choose interventions to solve the remaining causal structure learning problem of identifying whether batteries are good (causes of the LED's illumination) or bad (non-causes).

In order to disambiguate between causal structures in a way that is informative from an information-theoretic perspective, interventions should be chosen with the goal of reducing uncertainty about causal structures. This uncertainty reduction is also called information gain \citep{oaksfordRationalAnalysisSelection1994}.
Maximizing information gain corresponds to choosing interventions that can quickly narrow down which beliefs are most likely correct. In our LED example, if the child only has two batteries to test, then they are trying to learn which causal structure is correct among four possibilities: neither battery is good, only the first is good, only the second is good, or both are good.
Intervening on a single battery would be informative because this intervention eliminates half of the possibilities at once: If the LED illuminates, the child can rule out half of the structures where the intervened battery is \textit{not} good, and if the LED does not illuminate, then the child can rule out the other half. The alternative of intervening on both batteries would only be informative if the child strongly expects neither battery is good: The LED would likely remain unlit, eliminating all three other possibilities. However, if the child instead expects at least one battery is good, then intervening on both would no longer be informative: The LED would likely illuminate, eliminating only the single possibility of neither battery being good.

Past models \citep[e.g.,][]{steyversInferringCausalNetworks2003,bramleyConservativeForgetfulScholars2015,coenenStrategiesInterveneCausal2015} found that maximizing information gain about causal structures produced good predictions of people's interventions. However, they made an important simplifying assumption: they only represented people's beliefs about causal structure within a single, isolated causal learning problem. These models do not predict that people actively ``learn to learn'', improving their interventions for learning new causal relationships in the future.
In order to formalize such behavior, it is critical to represent \textit{causal overhypotheses}---abstract beliefs about causal relationships that span multiple situations and constrain how we learn the specifics in each situation \citep{goodmanFactFictionForecast1955, kempLearningOverhypothesesHierarchical2007,lucasWhenChildrenAre2014,simLearningHigherorderGeneralizations2017}.
If people's interventions help them update causal overhypotheses in past learning problems, then in a new problem, they would not need to start from scratch but can choose interventions that are guided by these overhypotheses. In this way, interventions from one situation are able to influence interventions in another, allowing them to adapt and improve not just within the current situation but also across to future ones.

To accommodate causal overhypotheses, we propose a hierarchical Bayesian model that represents beliefs at multiple levels of abstraction, including both lower-level beliefs about the current causal relationship and higher-level overhypotheses about the general properties of causal relationships.
Like previous rational~\citep{andersonAdaptiveCharacterThought1990} models of active learning~\citep[e.g.,][]{oaksfordRationalAnalysisSelection1994,steyversBayesianAnalysisHuman2009,bramleyConservativeForgetfulScholars2015,coenenStrategiesInterveneCausal2015}, we frame causal learning in Bayesian terms and take the active learner's goal to be finding interventions that maximize information gain, with one key difference: We posit that learners have overhypotheses that they update in light of new evidence, and seek information not just about the causal relationship at hand, but also about these overhypotheses.

While our general approach requires only a well-defined probabilistic model that includes overhypotheses, we focus here on overhypotheses about the \textit{functional form} of causal relationships. The functional form governs how causes combine or interact to produce an effect (Fig.~\ref{fig:causal_graph}), e.g., do relationships tend to be deterministic or stochastic? Are multiple causes necessary to bring about an effect? We choose this focus because (1) it allows us to build on simple hierarchical models that give good accounts of human learning in the absence of an active learning element \citep{lucasLearningFormCausal2010}; and because (2) varying overhypotheses in this setting leads to clear and systematic differences in what interventions are more informative, whereas we would expect subtler effects in an experiment based on other salient studies of overhypotheses \citep[e.g.,][]{kempLearningOverhypothesesHierarchical2007,austerweilLearningHowGeneralize2019}, owing to fewer degrees of freedom in possible interventions, and the possibility of greater individual variability in prior beliefs.  %

To give an example of the functional form and how it may affect future intervention choices, consider our LED example again, but now suppose that we do not know that the LED will illuminate if one or more charged batteries are present. If we assume that good batteries (causes) are interchangeable, then we can express the functional form in terms of the voltage threshold for illuminating the LED (effect), or---in terms of our original variables---the number of good batteries that are required to make it illuminate.
One overhypothesis is that only a single good battery is necessary (\textit{disjunctive} functional form); another is that we need two or more batteries (\textit{conjunctive} functional form). Our overhypotheses might also capture how reliable we expect the illumination to be for a particular set of batteries: It might be \textit{deterministic} if our voltage threshold is exceeded by any amount, or it might be unreliable (or \textit{noisy}) if our threshold is barely exceeded.
A priori, we might expect a deterministic and disjunctive form \citep{lucasLearningFormCausal2010,luBayesianGenericPriors2008,mayrhoferSufficiencyNecessityAssumptions2016,schulzGodDoesNot2006}, but if we find we need more than one charged battery, we can update our overhypotheses about how LEDs work in general.
By transferring these overhypotheses, we can pick more informative interventions in future situations with new LEDs and batteries. For example, under disjunctive-favoring overhypotheses, intervening on singleton batteries would be informative, revealing a good battery whenever there is an LED illumination. However, under conjunctive-favoring overhypotheses, this strategy would not be informative at all. Conjunctive overhypotheses expect that no single battery, good or bad, is sufficient to illuminate the LED, so intervening on a single battery would result in an unlit LED. This outcome provides no information about whether that battery is good (but just not sufficient by itself) or bad. Instead, having conjunctive overhypotheses leads us to a better strategy of testing two or more batteries at a time. Now it is possible to cause LED illuminations that tell us our intervention contains at least two good batteries.

Since past models of active causal learning have largely focused on learning causal structure \citep[e.g.,][]{bramleyConservativeForgetfulScholars2015,coenenStrategiesInterveneCausal2015,steyversInferringCausalNetworks2003}, they have tended to assume the functional form was known in advance to experimental participants, or that the functional form was consistent with the simple expectation that a single cause was sufficient to produce or prevent an effect.
This assumption of causal sufficiency 
holds for a wide variety of phenomena in causal inference and appears to be a default expectation people have in unfamiliar contexts \citep{chengCovariationCausationCausal1997,gopnikDetectingBlicketsHow2000,tenenbaumStructureLearningHuman2001,griffithsStructureStrengthCausal2005, griffithsBayesBlicketsEffects2011,luBayesianGenericPriors2008},
but it is not always appropriate. Both children and adults can adjust their overhypotheses to learn other functional forms, where multiple causes may be needed to produce the effect (e.g., the conjunctive form), and they are able to transfer these overhypotheses to guide their causal inferences in new tasks \citep{lucasLearningFormCausal2010,lucasWhenChildrenAre2014,kosoyLearningCausalOverhypotheses2022, griffithsTheorybasedCausalInduction2009,luBayesianTheorySequential2016}.
In our hierarchical Bayesian model, we accommodate uncertainty in peoples' overhypotheses about the functional form. In situations where people might be expected to have very strong a priori expectations about the functional form, our model is essentially equivalent to \citeauthor{steyversInferringCausalNetworks2003}'s  (\citeyear{steyversInferringCausalNetworks2003}) Rational Identification Model and \citeauthor{bramleyConservativeForgetfulScholars2015}'s (\citeyear{bramleyConservativeForgetfulScholars2015}) Scholar Model. In other situations where the form is not known in advance and when many forms are possible, our model makes substantially different predictions.

\begin{figure}[t]
  \centering
  \includegraphics[width=\textwidth]{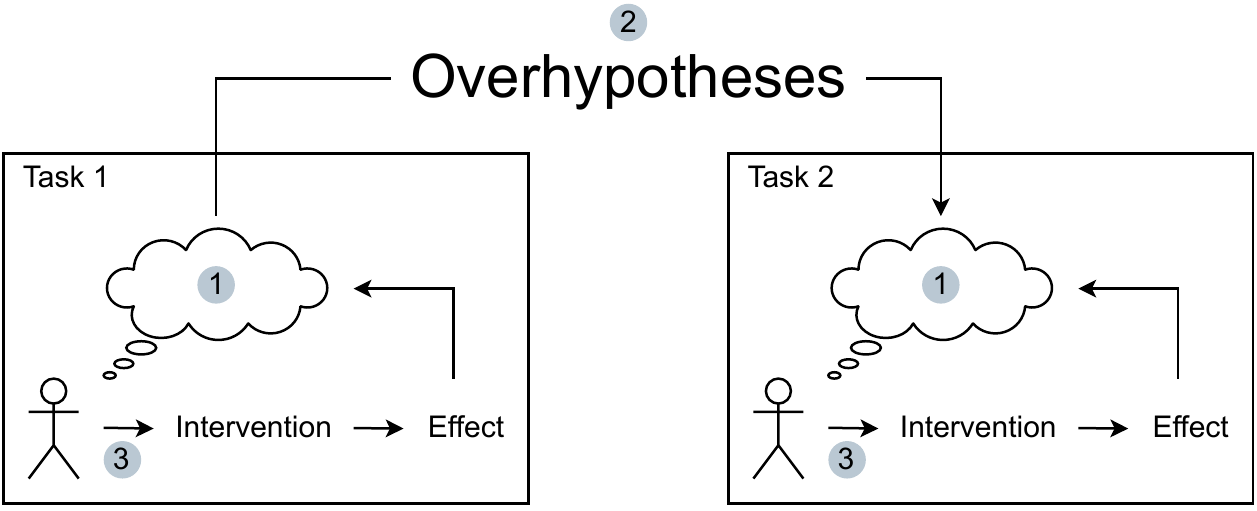}
  \caption{Hierarchical Bayesian model. Each circled number represents one of the three commitments of our model: (1) people represent a rich space of overhypotheses; (2) people transfer their learned overhypotheses from one task (left box) to the next (right box); and (3) they choose interventions that are informative for learning overhypotheses.
  }
  \label{fig:model_parts}
\end{figure}

Our model makes three commitments: (1) people represent a rich space of overhypotheses; (2) people transfer their learned overhypotheses from one task to the next; and (3) they choose interventions that are informative for learning overhypotheses (Fig.~\ref{fig:model_parts}). By removing the three modeling commitments one at a time, we create three ablation models for testing each commitment against alternative explanations (see Methods for the implementation of each commitment and ablation).

Our model's first commitment posits that people represent a rich space of overhypotheses. Following \citeauthor{lucasLearningFormCausal2010} (\citeyear{lucasLearningFormCausal2010}), we consider a \textit{sigmoid} space of overhypotheses about functional forms, where the common disjunctive and conjunctive forms are special cases within this space. The sigmoid space is computationally simple but is able to express rich variations in the number of causes required to generate an effect and in the reliability of the effect (see Methods).
To test this space, we compare our full model to a ``Fixed-Form'' ablation model that reduces the space of overhypotheses to a single deterministic and disjunctive form.
This ablation corresponds to the assumption that a single cause is sufficient to produce an effect, which is prevalent in past models of active causal learning \citep[e.g.,][]{bramleyConservativeForgetfulScholars2015,coenenStrategiesInterveneCausal2015,steyversInferringCausalNetworks2003}, and is also closely related to a \textit{positive testing} strategy.
In a positive testing strategy, interventions only test whether a causal relationship adheres to a single hypothesis, and it is possible that people mainly employ this strategy when they have a resource constraint \citep{coenenStrategiesInterveneCausal2015}, such as a time limit or a limit on the number of interventions.
This positive testing strategy may combine with people's prior preference for disjunctive and deterministic overhypotheses \citep{lucasLearningFormCausal2010,luBayesianGenericPriors2008,mayrhoferSufficiencyNecessityAssumptions2016,schulzGodDoesNot2006}, where a single cause is sufficient and reliable for producing the effect.
The resulting interventions would only test one object at a time, anticipating that a single object would be sufficient and reliable for producing the effect. Such interventions would not consider alternative overhypotheses like the conjunctive one, where at least two causes are needed to produce the effect and so is only revealed by testing combinations of objects.
This behavior constitutes an alternative to the commitment that people represent a rich space of overhypotheses.

Our model's second commitment posits that people transfer their learned overhypotheses from one task to the next. This means that when people encounter a new task, they do not start anew but instead ``learn to learn'', applying their previously learned overhypotheses to help them choose more efficient interventions for learning in the new task.
To test this transfer commitment, we compare our full model to the ``No-Transfer'' ablation model that predicts people start anew in each task using the same prior, regardless of any learning in previous tasks.
This ablation captures how people may have strong priors, specifically ones that favor deterministic and disjunctive overhypotheses \citep{lucasLearningFormCausal2010,luBayesianGenericPriors2008,mayrhoferSufficiencyNecessityAssumptions2016,schulzGodDoesNot2006}, and treat any learning about alternative overhypotheses as rare. They may then think these alternative overhypotheses are unlikely to be useful again, so instead of transferring these overhypotheses, they rely on the same deterministic- and disjunctive-favoring prior in a new situation.
This behavior constitutes an alternative to the commitment that people transfer their learned overhypotheses.

Our model's third and final commitment posits that people choose interventions that are informative for learning overhypotheses. Like previous models of active causal learning \cite[e.g.,][]{bramleyConservativeForgetfulScholars2015,coenenStrategiesInterveneCausal2015,steyversInferringCausalNetworks2003}, our model predicts that people seek information about the causal structure at hand. A key difference of our model lies in predicting that people also seek information about overhypotheses and that, following from our model's transfer commitment, people intend to transfer this information to future tasks. Our model captures the idea that people strike a balance between short-term learning, which deals with the causal structure at hand, and longer-term learning, which focuses on overhypotheses and how these can facilitate more informative interventions for the future.
To test our overhypothesis-learning commitment, we compare our full model to a ``Structure-Only-EIG (Expected Information Gain)'' ablation model that is short-sighted and only seeks information about the causal structure at hand, analogous to previous models of active causal learning. Any learning about overhypotheses and future benefits would only be \textit{incidental}, rather than a deliberate choice to pick interventions with an eye toward overhypotheses and future learning.
This behavior constitutes an alternative to the commitment that people choose interventions that are informative for learning overhypotheses.

Our full hierarchical Bayesian model was tested in two ways in our experiments: whether it was consistent with qualitative trends in our participants' interventions and judgments (Experiment 1), and whether it was the best model in a comparison against the ablation models and a random baseline, where models were ranked by how well they predicted participant interventions (Experiment 2).

\begin{figure}[t]
  \centering
  \includegraphics[width=0.84\textwidth]{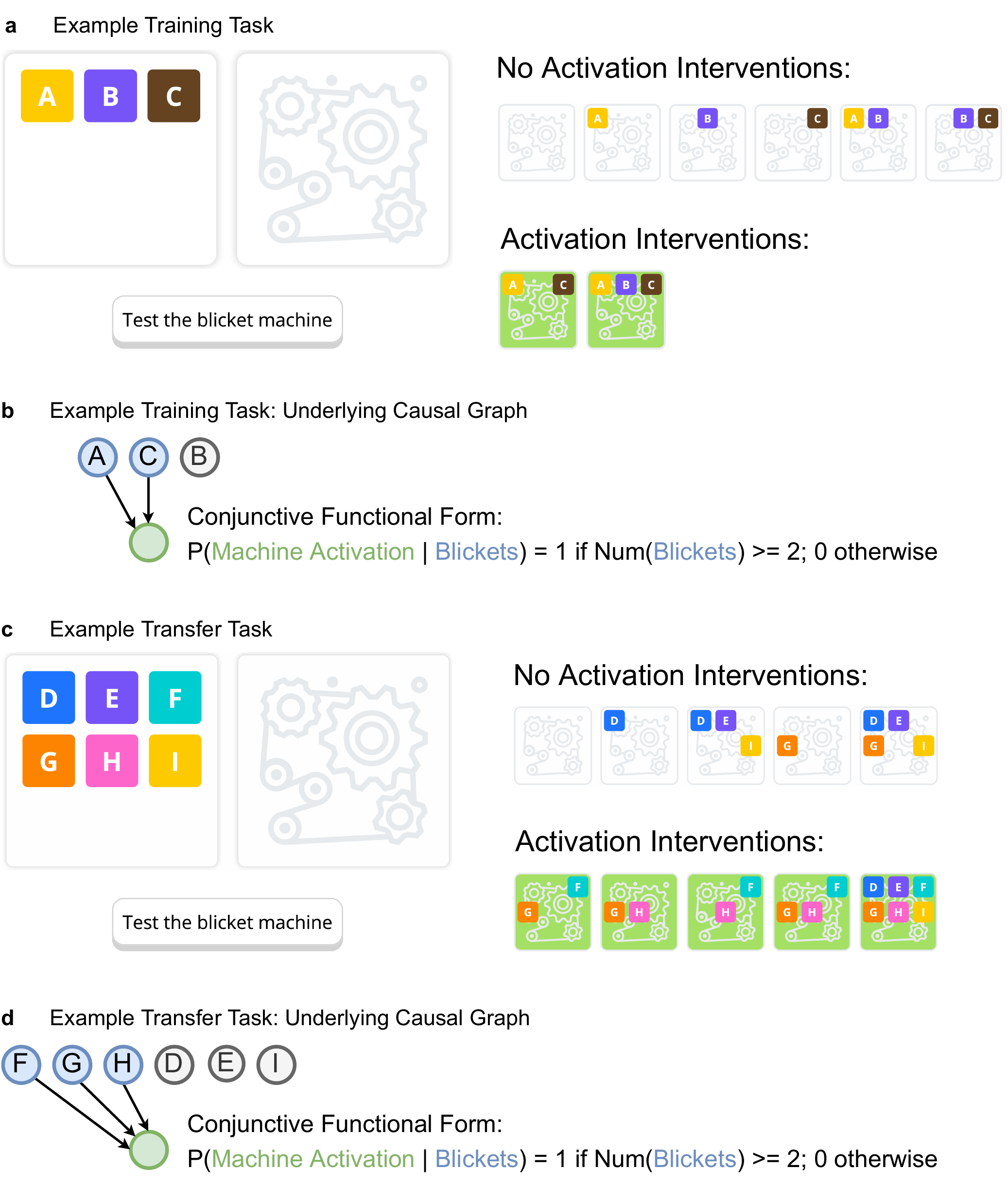}
  \caption{Example training and transfer tasks. \textbf{a}, Web interface for an example training task with 3 blocks (squares with colors and letters) and a blicket machine (embellished with cogs). An intervention involves clicking on blocks to set any combination on the machine and then pressing a button to test the machine's response (activation with a green color or nothing). Interventions must always contain A and C to activate the machine (shown to the right). \textbf{b} Causal graph of the example training task: The causal structure defines A and C as blickets (causes). The functional form is conjunctive and defines the conditional probability of the machine's activation (effect). \textbf{c} Web interface for an example transfer task with 6 blocks. Interventions must contain at least two of the blocks F, G and H to activate the machine (shown to the right; interventions are not comprehensive). \textbf{d} Causal graph of the example transfer task.}
  \label{fig:tasks}
\end{figure}

Our experiments were designed as active learning extensions of \citeauthor{lucasLearningFormCausal2010}'s (\citeyear{lucasLearningFormCausal2010}) version of the ``blicket detector'' experiments. As in \citeauthor{lucasLearningFormCausal2010}'s (\citeyear{lucasLearningFormCausal2010}) experiments, we presented participants with a task containing blocks (colored squares labeled with letters) and a ``blicket machine''. We asked them to solve the causal learning problem of identifying ``blickets'' (causes) among the blocks (prospective causes) by observing the blicket machine's binary response (effect). Whereas \citeauthor{lucasLearningFormCausal2010}'s study involved fixed sequences of events, ours used a computer-based web interface that allowed participants to actively produce their own sequences of events by choosing interventions (Fig.~\ref{fig:tasks}). An intervention involved putting any combination of blocks on the machine. The machine would then respond by activating or doing nothing. In order to choose informative interventions in this active blicket task, participants needed to consider their beliefs about both the causal structure (blicket identities) and the functional form (how the blicket machine activates in response to blickets). Participants encountered several active blicket tasks, where each one increased the level of difficulty and required increasingly selective interventions (see Methods).

We chose our active blicket experiment design because it (1) was simple enough for online experimental participants to quickly understand, (2) was tractable to analyze with our hierarchical Bayesian model, (3) nonetheless required appropriate overhypotheses about the functional form to facilitate learning in future tasks, and (4) could be decomposed into a causal structure learning aspect and a functional form learning aspect.
Within this design, we can formulate and test the ideas in our model: Participants’ interventions should not only learn the causal structure within each task, but they should also learn and transfer overhypotheses about the functional form to enable more efficient learning in future tasks.

\section{Results}\label{sec:results}

\subsection{Experiment 1}\label{sec:exp1}

Our first preregistered experiment is based on our conference paper that appeared in the CogSci Proceedings \citep{jiangExploringCausalOverhypotheses2021}. Here we tested our model's ideas based on qualitative trends in participants' interventions and causal judgments. Our model posits people choose interventions that learn overhypotheses about the functional form and thus enable them to learn more efficiently in future tasks. Following this, we hypothesized that in a new task (called the \textit{transfer} task), people would choose more efficient interventions and make more accurate judgments after training with the same functional form in past tasks. Conversely, they would choose less efficient interventions and make less accurate judgments after training with a different form. We also predicted that these effects would be larger if the same or different form was reinforced with more training tasks.

\begin{figure}[t]
  \centering
  \includegraphics[width=\textwidth]{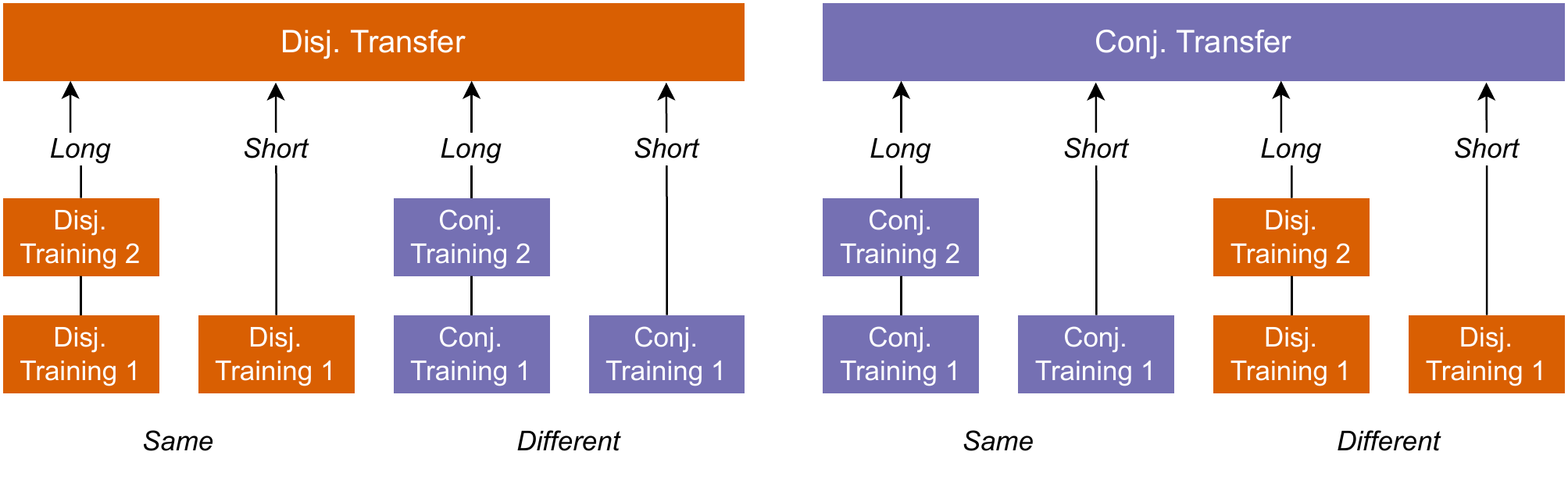}
  \caption{Experiment 1 conditions. Each of the 8 arrows represents a between-subjects condition and each box represents a training or transfer task. ``Disj.'' is short for Disjunctive and ``Conj.'' is short for Conjunctive. We manipulated the functional form of the transfer task (disjunctive or conjunctive), the training length (long with 2 training tasks, or short with 1 training task), and whether the training form was matched with the transfer form (same or different).}
  \label{fig:exp1_conditions}
\end{figure}

To test our hypothesis, we measured participants' ($N=\totalN{}$) interventions and causal judgments in the transfer task. We created 8 between-subjects conditions by manipulating three variables: the functional form of the transfer task (disjunctive or conjunctive), whether this form was matched with their past training tasks (same or different), and training length (short or long, i.e., one or two training tasks before the final transfer task) (Fig.~\ref{fig:exp1_conditions}; see Methods).

To represent the data accurately while accounting for potential data quality issues, we report results for both the full data set ($N=\totalN{}$) and a filtered subset ($N_f=\filteredN{}$), which includes most (\filterToTotalPercent{}) of the original participants while requiring more participant engagement. The filtered participants made at least 9 interventions in the transfer task, which was the minimum number required to execute a straightforward strategy in the easier disjunctive variant of the task: testing whether each of the 9 blocks was a blicket that can individually activate the machine.

\subsubsection{Causal judgments}
First we tested our hypothesis on participants' causal judgments in the transfer task, where they aimed to identify which of the 9 blocks were blickets.
Participants also judged whether the blicket machine would activate in response to different combinations of blocks. These judgments are described in the Supplementary Results, and were broadly consistent with blicket identification judgments. Here we focus on the latter because the blicket identification format follows more closely from past blicket studies \citep{lucasLearningFormCausal2010}.

We expected causal judgment accuracy to be predicted by the match between the transfer and training functional forms, training length, and their interaction, considering the transfer task's functional form as a covariate.
We used these variables to fit a logistic regression model to predict the per-participant accuracy percentage in the transfer task's blicket identification questions (binomial with 9 trials).
We confirmed a significant main effect of the match of functional form ($z=2.62, p=.009$; filtered: $z_f=3.25, p_f=.001$).
The transfer task's functional form also had a significant main effect ($z=3.99, p<.001$; filtered: $z_f=4.78, p_f<.001$), which was consistent with past results suggesting that people find disjunctive forms easier to learn \citep{lucasLearningFormCausal2010}.
Surprisingly, the length of training and its interaction with the match between training and transfer forms were not significant predictors. 

We also used Welch t-tests (two-tailed) to investigate the specific effects of match between pairs of conditions (visualized in Supplementary Results Fig.~\ref{fig:exp1_quiz}a), expecting causal judgment accuracies to improve from mismatched to matched conditions.
In the disjunctive transfer conditions, the comparisons were mostly consistent with our expectations: in the full data, the mean blicket identification accuracy showed a trend toward improvement from mismatched (conjunctive training) to matched (disjunctive training) conditions with long, $t(47.82)=-2.00, p=.051$, and short training, $t(49.76)=-1.80, p=.078$. These trends became significant improvements in the filtered data where participants were more engaged (long: $t_f(42.10)=-2.18, p_f=.035$; short: $t_f(43.99)=-2.47, p_f=.017$).
In the conjunctive transfer conditions, however, the difference between matched (conjunctive training) and mismatched (disjunctive training) accuracies was non-significant. This weaker match effect might have accounted for the non-significant interaction effect between match and training length in our logistic regression model. 
We suspected this weaker effect was due to the conjunctive transfer task being too difficult to learn, regardless of training match and length. This suspicion was supported by the results in our next experiment, where we lowered the difficulty of the conjunctive transfer task and found a significant improvement from mismatched to matched conditions (see Supplementary Results).

\subsubsection{First intervention}
Then we tested our hypothesis on participants' \textit{first} intervention in the transfer task. The first intervention had to be chosen before participants learned anything about the functional form in the transfer task, making it a simple marker of how participants' interventions were informative under a functional form from their past training. Under a disjunctive form, intervening on a single block would be informative for identifying blickets, requiring only nine interventions in all. In contrast, this would be completely uninformative under a conjunctive form, which would require intervening on more blocks at a time to identify blickets.
We expected that participants' first intervention in the transfer task would be informative under their training form, and therefore, we also expected this intervention to be more efficient if the training and transfer forms were the same.

\begin{figure}[t]
  \centering
  \includegraphics[width=\textwidth]{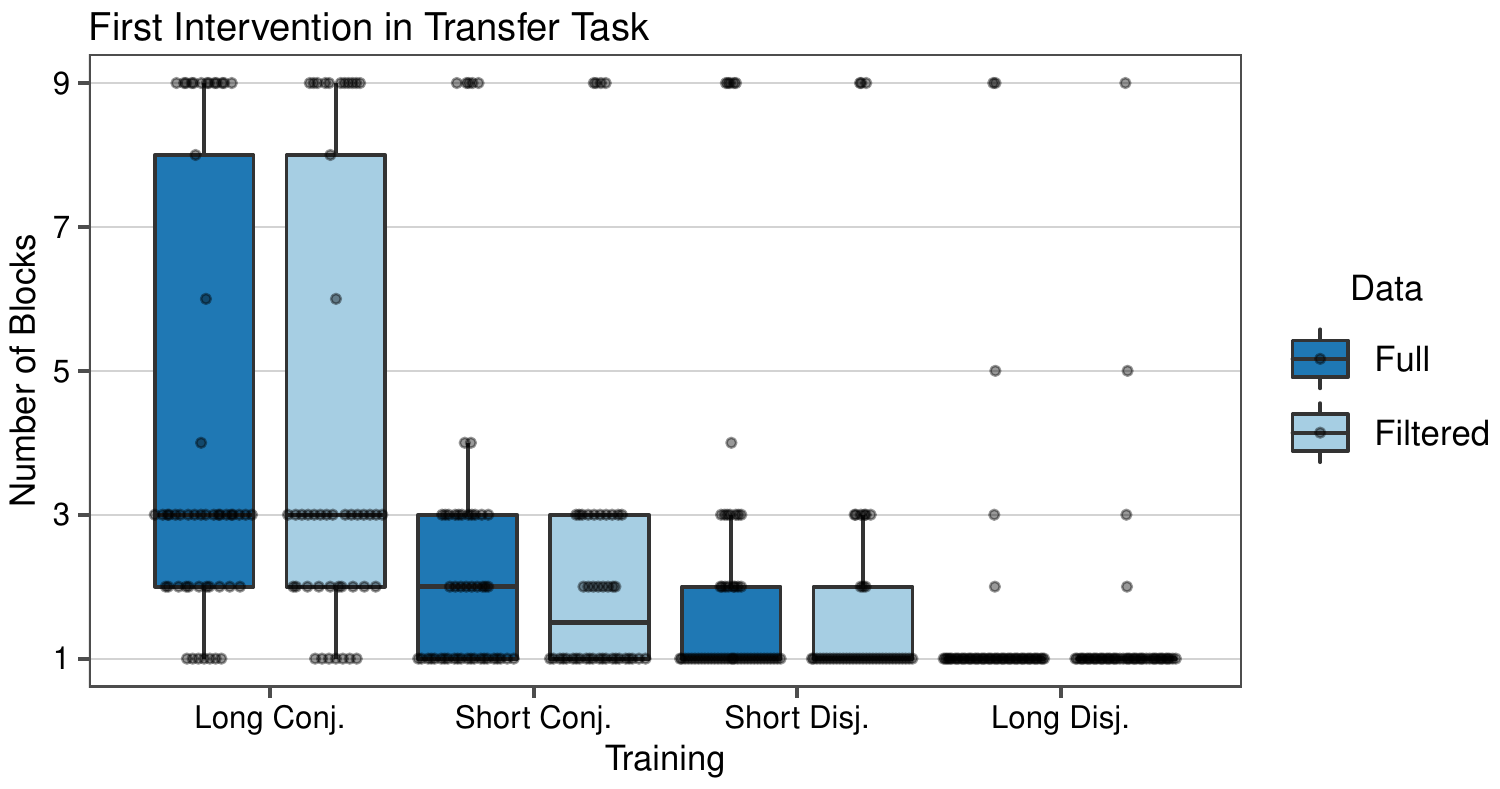}
  \caption{Experiment 1: Number of blocks in the first intervention in the transfer task. This is plotted against the functional form of past training tasks and the training length. The box-and-whisker plots show the quartiles of the full or filtered data; each overlaid point represents a participant.}
  \label{fig:exp1_numBlocks_training}
\end{figure}

To test whether the first intervention in the transfer task was informative under the training form, we used a linear model to predict the number of blocks in the first intervention, where the predictors were the training form, the training length, and their interaction. There was a significant interaction effect ($t(205)=-3.59, p<.001$; filtered: $t_f(177)=-3.46, p_f<.001$) and significant main effect of training length ($t(205)=3.38, p<.001$; filtered: $t_f(177)=3.66, p_f<.001$). The non-significant main effect of the training form ($p \geq .322$ for both the full and filtered data) may be attributable to weaker effects in the short conditions---see Fig.~\ref{fig:exp1_numBlocks_training}.
Consistent with these model results and with the trends in Fig.~\ref{fig:exp1_numBlocks_training}, the mean number of blocks in the first intervention was significantly higher after conjunctive training (short and long) than after disjunctive training (short and long), $t(183.39)=4.62, p<.001$ (filtered: $t_f(144.96)=4.75, p_f<.001$), suggesting participants' interventions were informative under their training form. 

\begin{figure}[t]
  \centering
  \includegraphics[width=\textwidth]{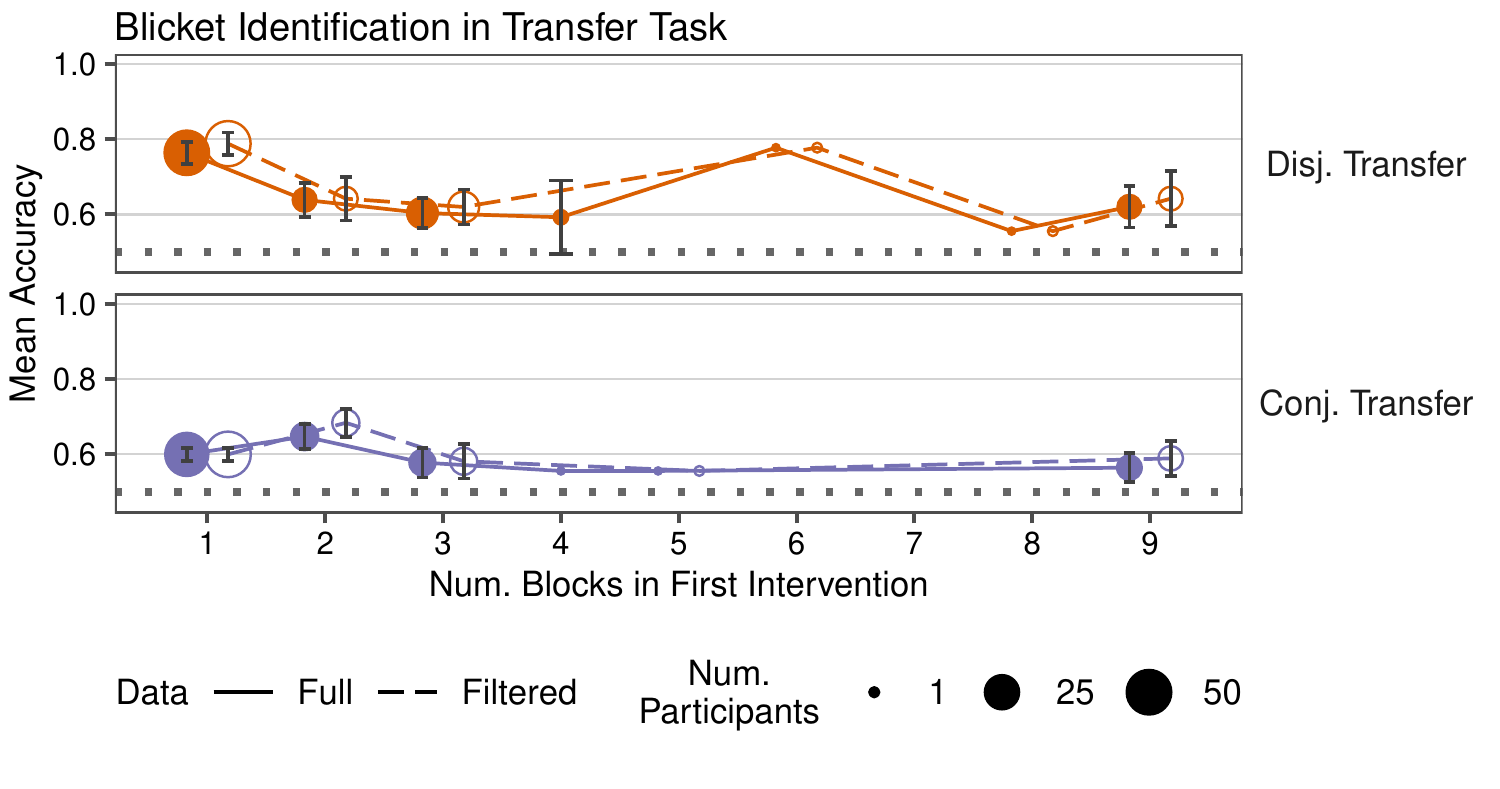}
  \caption{Experiment 1: Mean participant accuracies for blicket identification questions in the transfer task. This is grouped by the number of blocks in the first intervention and the transfer form. For a particular number of blocks, the size of the dot represents how many participants are involved in calculating the mean accuracy. The mean is calculated separately for the full (solid lines) and filtered (dashed lines) data. Chance (.5) accuracy is shown with a dotted gray line. Error bars in either direction denote the magnitude of the standard error but are omitted for points with a single participant, where the standard error is ill-defined.}
  \label{fig:exp1_blicketAcc_numBlocks}
\end{figure}

To understand when the first intervention would be efficient for learning in the transfer task, we fitted a (binomial) logistic regression model to predict blicket identification accuracy in the transfer task (9 trials). The predictors included the number of blocks in the first intervention, the functional form of the transfer task, and their interaction. There was a significant main effect of the transfer form ($z=4.26, p<.001$; filtered: $z_f=4.90, p_f<.001$), underscoring the relative difficulty of the conjunctive condition, and no significant main effect for the number of blocks (all $p \geq .382$), suggesting that any effect of the number of blocks was not due to choosing more (or fewer) blocks being a better general-purpose policy. Rather, the effect of the number of blocks was due to being informative of a particular transfer form: this interaction did not reach significance in the full data ($z=-1.73, p=.084$), but was significant for the more engaged participants in the filtered data ($z_f=-2.02, p_f=.043$).
Specifically, Fig.~\ref{fig:exp1_blicketAcc_numBlocks} shows the disjunctive accuracy peaks at a singleton block and has a decreasing trend as the number of blocks increases, while the conjunctive accuracy peaks at two blocks.

Putting together the results on the first transfer intervention, participants chose interventions that tended to be informative under their training form: they chose fewer blocks after disjunctive training than after conjunctive training, with a focus on singleton blocks after disjunctive training (Fig.~\ref{fig:exp1_numBlocks_training}'s median lines). This first intervention was consistent with more efficient learning in a matched transfer task: a singleton block indicated better blicket identification in the disjunctive transfer task and two blocks indicated better blicket identification in the conjunctive transfer task. Combined with the results on participants' causal judgments, Experiment 1 supports our hypothesis that, in the new transfer task, people choose more efficient interventions and make more accurate causal judgments after training with the same functional form.
These trends suggest people learn and transfer overhypotheses about the functional form, and they are able to exploit these overhypotheses to improve their active learning in similar future situations. Thus, Experiment 1 provides qualitative evidence for the ideas in our hierarchical Bayesian model.

\subsection{Experiment 2}\label{sec:exp2}

In our second preregistered experiment ($N = \twoTotalN{}$), we continued to test our hierarchical Bayesian model in a formal model comparison where models were ranked by how well they predicted participant interventions. By comparing our model against ablation models, we could test each of our model's commitments: (1) people represent a rich space of overhypotheses; (2) people transfer their learned overhypotheses from one task to the next; and (3) they choose interventions that are informative for learning overhypotheses (Fig.~\ref{fig:model_parts}).

\begin{figure}[t]
  \centering
  \includegraphics[width=\textwidth]{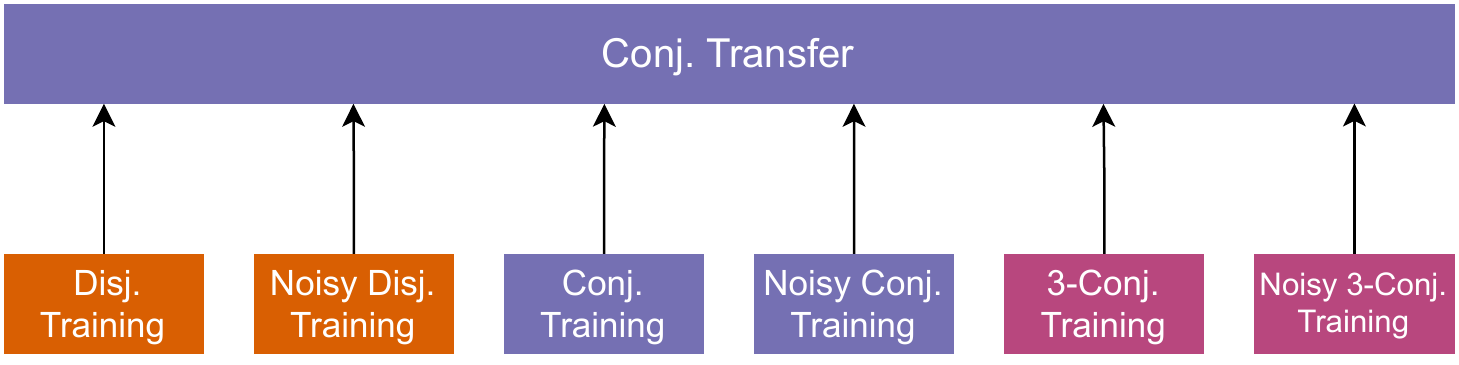}
  \caption{Experiment 2 conditions. Each of the 6 arrows represents a between-subjects condition and each box represents a training or transfer task. ``Disj.'' is short for Disjunctive, ``Conj.'' is short for Conjunctive, and ``3-Conj.'' is short for 3-Conjunctive. We fixed the transfer task's form as conjunctive and manipulated the functional form of the training task. For the training form, we varied both the minimum number of blickets needed to activate the blicket machine---1 (disjunctive), 2 (conjunctive), or 3 (3-conjunctive)---and whether that activation was deterministic or noisy (probability 0.75 of activation given the minimum number of blickets).}
  \label{fig:exp2_conditions}
\end{figure}

Experiment 2 was designed to test our model's commitments: it expanded the manipulation of the functional form in the training task to be representative of the rich space considered by our model (commitment 1). We varied both the minimum number of blickets needed to activate the blicket machine---1 (disjunctive), 2 (conjunctive), or 3 (3-conjunctive)---and whether that activation was deterministic or noisy (probability 0.75 of activation given the minimum number of blickets), creating 6 between-subjects conditions (the training length was fixed to a single task; see Fig.~\ref{fig:exp2_conditions} and Methods). The transfer task was fixed to have the same deterministic conjunctive form for all 6 between-subjects conditions, which tested how interventions in the same task would differ depending on overhypotheses learned from past training tasks. No matter if the training form was a mismatched (noisy) disjunctive form, a matched (noisy) conjunctive form, or a similar (noisy) 3-conjunctive form, we wanted our model to capture how participants transferred overhypotheses about the training forms (commitment 2). Finally, participants needed to choose informative interventions about overhypotheses to successfully find blickets in the transfer task (commitment 3).

\subsubsection{Model comparison}
We evaluated our model by how well it predicted participant interventions in the transfer task, calculated as the predictive likelihood (see Methods). If our model assigned a high predictive likelihood to participant interventions, then there was a good alignment between model predictions and participant interventions. For each participant intervention, there were $2^6=64$ possible combinations of the 6 transfer task blocks that the participant could intervene on. This sets the random baseline predictive likelihood at $1/64$. To perform better than this random baseline, our model needed to make specific predictions about which of the 64 possible interventions were good, and this prediction needed to align with the participant's single chosen intervention.

\begin{figure}
\centering
\includegraphics[width=0.9\textwidth]{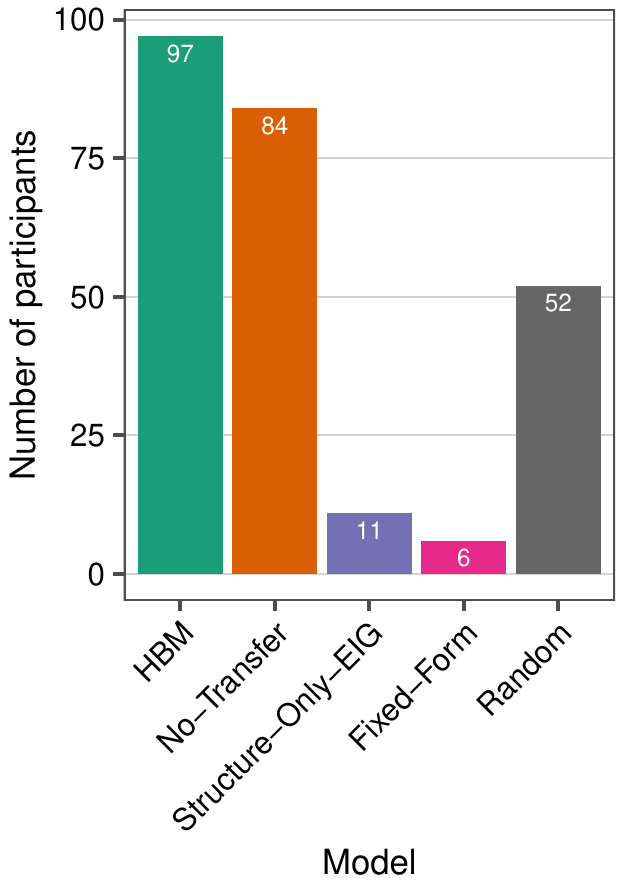}
\caption{Best model for individual participants (Experiment 2). Each model's bar counts the number of participants whose transfer task interventions were best-predicted by that model, i.e., assigned the highest mean predictive likelihood by that model compared with all other models. The comparison was performed on a per participant basis and using cross-validation and marginalization over priors (see Methods). Our Hierarchical Bayesian Model (HBM) was the best predictor for the highest number of participants compared with the ablation models (No-Transfer, Structure-Only-EIG (Expected Information Gain), Fixed-Form) and a random baseline.}
\label{fig:model_comparison_ind}
\end{figure}

We compared our model's predictive likelihoods to the random baseline as well as the ablation models (using cross-validation and marginalization over priors; see Methods). In an averaged comparison across all participants and their interventions, the No-Transfer ablation model had the highest mean predictive likelihood ($\text{mean (M)} \pm \text{standard error (s.e.)} = .0581 \pm  .0022$) and was closely followed by our full model ($\text{M} \pm \text{s.e.} = .0529 \pm  .0024$). All other models had lower predictive likelihoods (Fixed-Form: $\text{M} \pm \text{s.e.} = .0183 \pm  .0003$; Structure-Only-EIG: $\text{M} \pm \text{s.e.} = .0148 \pm  .0008$; Random (fixed value): $.0156$). In an individual differences analysis, we compared all models on a per-participant basis and distributed participants by their best  model. Here we found that No-Transfer was now the second-best model and that our full model was the best predictor for the highest number of participants (Fig.~\ref{fig:model_comparison_ind}).

\begin{figure}
\centering
\includegraphics[width=\textwidth]{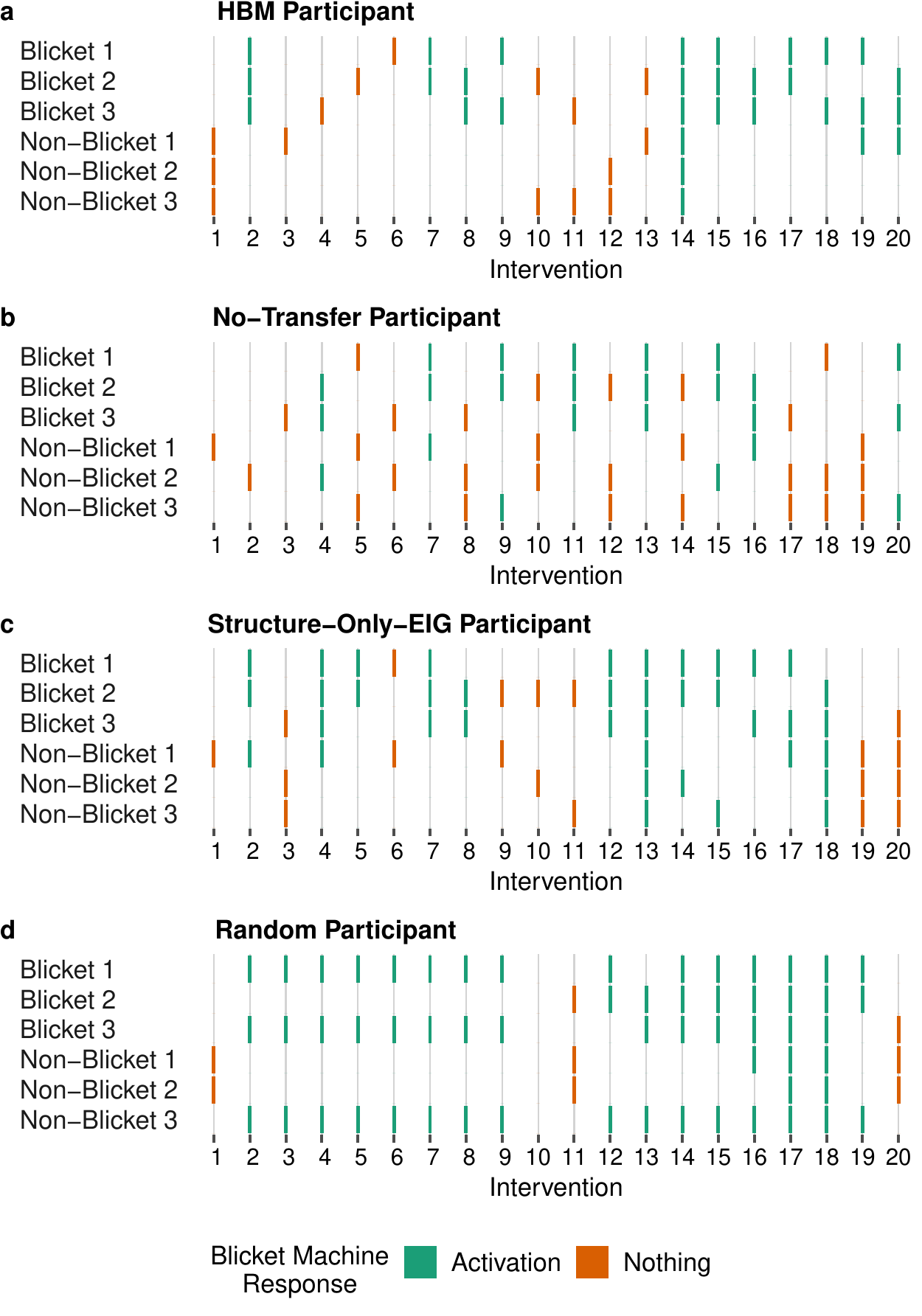}
\caption{Representative individual participants for each model (Experiment 2). Their 20 interventions are shown for the conjunctive transfer task and they had previously completed a matched conjunctive training task (i.e., they all belonged to the conjunctive training condition). Each intervention contains some combination of the transfer task's 6 blocks, whose identities (blicket or non-blicket) are labeled in the figure but were not known to participants. The blicket machine's response (activation or nothing) is marked with the color and was also known to participants. The Fixed-Form ablation model is not included because it was not the best predictor for any participant in the conjunctive training condition.}
\label{fig:example_ind}
\end{figure}

Fig.~\ref{fig:example_ind} illustrates a representative individual participant for each model. This figure shows participants from only the conjunctive training condition and their 20 interventions in the conjunctive transfer task (see Fig.~\ref{fig:tasks} for examples of these conjunctive training and transfer tasks). Looking at the same training condition makes it possible to tease apart individual differences that are captured by differences in the models rather than differences in participants' past training experience.

Our hierarchical Bayesian model (HBM) best-predicted a participant who immediately began the transfer task by testing multiple blocks (Fig.~\ref{fig:example_ind}a; interventions 1-2), suggesting they transferred conjunctive-favoring overhypotheses. Next, they tested singleton blocks (interventions 3-6) that were informative for disambiguating conjunctive forms from disjunctive ones within a rich space of overhypotheses. Within the next few interventions (7-9) they chose interventions that were also informative for learning causal structure, confirming which blocks were blickets under the conjunctive form. Thus, all of our model's commitments were crucial for predicting this participant's intervention strategy.

The ablation and random models captured other qualitatively distinct strategies.
The No-Transfer ablation model best-predicted a participant who began with a similar strategy to the HBM participant, except they prioritized singleton interventions first before intervening on multiple blocks (Fig.~\ref{fig:example_ind}b). This strategy suggests they initially prioritized learning disjunctive forms, which is consistent with the disjunctive-favoring priors that people tend to bring to the blicket experiment \citep{lucasLearningFormCausal2010}, rather than overhypotheses transferred from the previous conjunctive training task (i.e., ablating the transfer commitment).
The Structure-Only-EIG ablation model best-predicted a participant whose interventions were consistent with having learned conjunctive-favoring overhypotheses from training, transferring these overhypotheses to the transfer task, and then prioritizing learning causal structure under these overhypotheses: they focused heavily on multiple-block interventions, suggesting they were searching for blickets under conjunctive-favoring overhypotheses (Fig.~\ref{fig:example_ind}c).
They did not prioritize learning other possible forms (i.e., ablating the overhypothesis learning commitment), as indicated by choosing only one singleton intervention, which was not enough to detect other forms like the disjunctive form.
The Fixed-Form ablation model was not the best predictor for any participant in the conjunctive training condition, suggesting that, in the conjunctive training and transfer tasks, participants considered alternative forms beyond a single disjunctive functional form (i.e., keeping the commitment to representing a rich overhypothesis space).
Finally, the random baseline model best-predicted a participant whose interventions were not compatible with other models (Fig.~\ref{fig:example_ind}d). This participant appeared to employ a low-effort strategy where they were trying to complete the transfer task quickly (perhaps to receive their performance-independent compensation) rather than learn the task, a point which we cover in the Discussion.

Overall, participants employed several qualitatively different strategies in the transfer task, even when their previous training was the same (Fig.~\ref{fig:example_ind}). On average, these strategies were most consistent with the No-Transfer ablation model, but it is unlikely that this single model was representative of all the qualitatively different strategies employed by different individuals; we elaborate on this in the Discussion. To accommodate this variability, we performed an individual differences analysis to find the best model per participant. The largest group of individuals employed a strategy that was best-predicted by our hierarchical Bayesian model, suggesting our model captured people's dominant intervention strategy.

\section{Discussion}\label{sec:disc}

How do people actively learn to learn causal relationships? That is, how and when do people choose interventions that facilitate long-term learning and promise more-informative future interventions?
We proposed and tested a hierarchical Bayesian model that goes beyond past models of active causal learning by representing people's beliefs at multiple levels of abstraction. The lower-level beliefs are about the causal relationship in front of them, while the higher-level beliefs, called overhypotheses \citep{goodmanFactFictionForecast1955,kempLearningOverhypothesesHierarchical2007}, are about general causal properties that can be shared with future causal systems.
We focused on overhypotheses about the functional form, which governs how multiple causes combine or interact to produce an effect.
Our model rests on three testable commitments: (1) people represent a rich space of overhypotheses; (2) people transfer their learned overhypotheses from one task to the next; and (3) they choose interventions that are informative for learning overhypotheses.
We tested how well these commitments explained human behavior in our two active blicket experiments, where we used a total of 14 between-subjects manipulations of training and transfer tasks to probe how and when people choose interventions that facilitate long-term learning across those tasks.
Our model's commitments were first supported by qualitative trends (Experiment 1), where participants' interventions and judgments showed long-term improvement when the training and transfer tasks had the same functional form.
Our model was further supported by an individual-differences-based model comparison (Experiment 2) that demonstrated our model was the best predictor for the largest group of participants.
Our results suggest
when there are abstract similarities across active causal learning problems,
people readily learn and transfer overhypotheses about these similarities.
Moreover, people exploit these overhypotheses to facilitate long-term active learning.

Our hierarchical Bayesian model was the best predictor in our individual differences analysis, but it was only the second-best in an averaged analysis of all participants.
The No-Transfer ablation model was best on average, which could be because this model had a diffused prior in the transfer task. 
Since the No-Transfer model removes the commitment that people transfer what they learned from the previous training tasks, it would start anew with a prior that has not been tuned to any individual's previous learning. Such a prior could be reasonably consistent with several strategies that are informative about any of the beliefs that are likely under that diffused prior. For example, these strategies could include our full model's strategies, which are informative about transferred overhypotheses, or less cognitively demanding strategies, where even random strategies can update \textit{some} set of beliefs. Thus, the No-Transfer model would be, on average, a good predictor for all of these different strategies.
However, the average model does not explain how people can also deviate from the average and behave differently from each other. Moreover, it is often the case that the average model may not capture the behavior of any individual \citep[e.g.,][]{hayesBackwardCurveMethod1953,estesProblemInferenceCurves1956,ashbyDangersAveragingSubjects1994,heathcotePowerLawRepealed2000}, and this phenomenon is also pertinent to average models of causal learning \citep{johnstonIndividualDifferencesCausal2021}.

In our study, individuals employed several qualitatively distinct strategies (Fig.~\ref{fig:example_ind}) that were unlikely to be captured by an average model, so we conducted an individual differences analysis.
This diversity in strategies, including some that may result in lower judgment accuracy or less informative interventions while imposing lower demands on memory, time, or attention, is consistent with a ``resource-rational" view of inductive learning~\citep{griffithsRationalUseCognitive2015,liederStrategySelectionRational2017}.
For example, if a participant is not motivated to perform well in the task itself but is rather only concerned with completing the task and receiving the base compensation (independent of task performance), they might choose a cost-efficient strategy that is more consistent with a random or ablation model.
Under this view that individuals vary in their implicit cost/performance trade-offs, greater performance incentives (e.g., bonuses to our crowdsourced participants) may increase the proportion of participants who are best-predicted by our full hierarchical Bayesian model. Setting aside that conjecture, we found the largest group of participants chose a strategy that was best-predicted by our full model.

The No-Transfer ablation model also performed well in our individual differences analysis and captured a sizable, second-largest group of individuals.
These individuals may have had sufficiently strong prior beliefs favoring deterministic and disjunctive relationships \citep{lucasLearningFormCausal2010,luBayesianGenericPriors2008,mayrhoferSufficiencyNecessityAssumptions2016,schulzGodDoesNot2006} such that they treated the conjunctive and 3-conjunctive training conditions as outliers or special cases, unlikely to generalize to new machines.
Consequently, they may have resorted to the same deterministic- and disjunctive-favoring prior in the next task, as is consistent with the No-Transfer model. Such behavior is also consistent with how people generalize causal laws across several tasks \citep{zhaoHowPeopleGeneralize2022}.
The alternative behavior would be to expect a high degree of similarity in our experiment's training and transfer tasks and thus find it useful to transfer overhypotheses between these tasks, as is consistent with our full model.
Although the No-Transfer participants were not captured by our full model, we note that their behavior was still consistent with the other two commitments that were retained in the No-Transfer model: representing a rich space of overhypotheses and choosing interventions that are informative for learning overhypotheses. Both of these commitments go beyond past models of active causal learning and were required to predict the majority (72.4\%) of participants' interventions (combining both HBM and No-Transfer participants in Fig.~\ref{fig:model_comparison_ind}).
Ultimately, the largest group of participants was best-predicted by our full model, suggesting that all three commitments, including the transfer commitment, were required to capture the dominant strategy that people employ in actively learning to learn causal relationships.

Following \citet{lucasLearningFormCausal2010}, we chose our space of overhypotheses about functional forms by assuming (1) causes are interchangeable, without any distinct subtypes of causes; (2) as the number of active causes grows, so does the probability of the target effect; (3) causes may or may not be individually sufficient to generate the effect, and (4) relationships can (but need not) be deterministic. These assumptions are captured by letting the probability of the effect (blicket machine activation) be a sigmoid function of the number of causes (blickets). It captures several qualitatively distinct relationship types, including all of the forms in our experimental manipulations, while being computationally efficient, interpretable, and having only two parameters (see Methods).
While this space is suited to our blicket cover story and simplifies model-building, people are able to learn and transfer a broader space of forms than can be encompassed by the sigmoid space, such as forms that combine preventative and continuous causes \citep{yuilleNoisyLogicalDistributionIts2007,luBayesianTheorySequential2016}. Therefore, an exciting future direction is to go beyond parametric functional forms, and consider arbitrarily expressive belief spaces. For example, grammar-based and program induction methods offer suggestions about how people can dynamically and compositionally expand their belief space with an infinite set of possible concepts  \citep[e.g.,][]{goodmanRationalAnalysisRuleBased2008,lakeHumanlevelConceptLearning2015,goodmanConceptsProbabilisticLanguage2015,piantadosiLogicalPrimitivesThought2016}.

An alternative family of models for active learning originates in the reinforcement learning literature. Models based on reinforcement learning have successfully explained cognitive phenomena \citep[e.g.,][]{dayanReinforcementLearningGood2008} and provided computational solutions to complex active learning problems \citep[e.g.,][]{vinyalsGrandmasterLevelStarCraft2019,wurmanOutracingChampionGran2022}.
These models typically require thousands of actions and tasks where a human may only require a few, but recent advances have begun to leverage abstract knowledge that can be shared between tasks and thus allows a reinforcement learning model to learn more efficiently in new tasks \citep{hospedalesMetaLearningNeuralNetworks2020,tomovMultitaskReinforcementLearning2021,ecksteinComputationalEvidenceHierarchically2020,zhangLearningInvariantRepresentations2021}.
However, it is still a challenge for these models to incorporate certain kinds of abstract knowledge and inductive biases that align with human behavior, especially human causal learning.
For example, modern reinforcement learning agents have difficulty learning abstract causal knowledge in the Alchemy benchmark \citep{wangAlchemyBenchmarkAnalysis2021}, which was designed with inspiration from studies of human learning.
Furthermore, with regard to blicket tasks that are similar to our experiments, it remains an open direction how current reinforcement learning agents can explore like children \citep{kosoyLearningCausalOverhypotheses2022}, who consider rich priors over causal overhypotheses that are much like those studied in our current work.
This is not to say the reinforcement learning approach would be ineffective for modeling how humans actively learn to learn, but there are open questions about how this approach can achieve the same learning efficiency and inductive biases as humans.
Thus, in our work we have chosen a hierarchical Bayesian model that can learn from the same number of interventions as each participant (see Methods), and can straightforwardly represent and learn about overhypotheses that are supported by studies of human behavior \citep{lucasLearningFormCausal2010,lucasWhenChildrenAre2014}.

Overall, we explored the question of how humans choose actions to facilitate long-term learning and make their future actions more efficient. In other words, how do people actively learn to learn? We focused on the domain of active causal learning, where past models \citep[e.g.,][]{steyversInferringCausalNetworks2003,bramleyConservativeForgetfulScholars2015,coenenStrategiesInterveneCausal2015} have made an important simplifying assumption by predicting that interventions are only informative about the causal relationship at hand, which would not explain how people can learn beyond their current situation and choose more efficient interventions in the future.
We proposed and found evidence for a hierarchical Bayesian model, which differs from these earlier models in one key way: it posits that people not only seek information about the causal relationship at hand but also pick interventions with the goal of learning and transferring \textit{overhypotheses} \citep{goodmanFactFictionForecast1955,kempLearningOverhypothesesHierarchical2007} that are useful for future causal learning problems.
Our approach generalizes beyond causal learning to active learning in any setting where there is an opportunity for learning about the abstract properties of the task, i.e., for updating and exploiting overhypotheses. Examples range from graph structure learning \citep{mansinghkaStructuredPriorsStructure2006} to the optimal stopping problem \citep{leeHierarchicalBayesianModel2006} to category learning \citep{kempLearningOverhypothesesHierarchical2007}.
Thus, accounting for overhypotheses may provide a foundation for working toward a better understanding of a wide range of domains where humans actively learn to learn.

\section{Methods}\label{sec:methods}

\subsection{Preregistrations}
Experiment 1's preregistration: \url{https://osf.io/n9cx2/?view_only=caa45bfd6c8c4d1ebf2c904878d3fff8}

Experiment 2 and hierarchical Bayesian model preregistration: \url{https://osf.io/vk9yd/?view_only=dd45fd1aafd9499d8173c64cae2deedc}

\subsection{Overall Experiment Design}

Each active blicket task can be formalized as learning a causal graph (see Fig.~\ref{fig:tasks} and Fig.~\ref{fig:causal_graph}). The \textit{causal structure} of the graph defines what variables are causes and effects of other variables. The variables in the task are the blocks' presence on the machine and the blicket machine's activation.
From the cover story, participants can easily identify the machine activation as the only plausible effect, but they do not know whether the other variables (blocks) are causes (blickets) or non-causes (non-blickets). Their goal is to solve the causal structure learning problem of identifying causes from non-causes, or blickets from non-blickets. 

To identify blickets, participants must, however, also solve the \textit{functional form} learning problem. The functional form defines the conditional probability of the effect (machine activation) given the causes that are present (blickets, not non-blickets, that are on the machine), and it is a function of these causes (blickets). For example, a disjunctive form says the conditional probability is 1 whenever at least one blicket is on the machine and 0 otherwise. A conjunctive form changes the blicket ``threshold'', saying the conditional probability of a machine activation is 1 when at least two blickets are on the machine and 0 otherwise; see Table~\ref{tab:fforms} for all the functional forms we consider.
Under a disjunctive form, participants can intervene on one block at a time to identify whether that block is a blicket that activates the machine. However, under a conjunctive form, singleton interventions would not reveal anything about blickets. Participants would instead need to intervene on multiple blocks at a time to reveal any machine activations, and from there on, they might still need to narrow down which blocks in their intervention are actually blickets.
Thus, in order to achieve the goal of identifying blickets, participants must learn both the causal structure and functional form of the task. 

\begin{table}[]
  \caption{Functional forms}
  \label{tab:fforms}
  \centering
  \begin{tabular}{@{}lllll@{}}
    \toprule
    \multirow{2}{*}{Functional form} & \multicolumn{2}{l}{Interpretation}                           & \multicolumn{2}{l}{Sigmoid param.}                                                 \\ \cmidrule(l){2-5} 
                                     & \begin{tabular}[c]{@{}l@{}}Blicket                                                                                                                \\ threshold\end{tabular} & \begin{tabular}[c]{@{}l@{}}Activation \\ probability\end{tabular} & Bias & Gain                          \\ \midrule
    Disjunctive         & 1 & 1   & 0.5 & $\gg$ 1 \\
    Noisy Disjunctive   & 1 & .75 & 0.9 & 11      \\
    Conjunctive         & 2 & 1   & 1.5 & $\gg$ 1 \\
    Noisy Conjunctive   & 2 & .75 & 1.9 & 11      \\
    3-Conjunctive       & 3 & 1   & 2.5 & $\gg$ 1 \\
    Noisy 3-Conjunctive & 3 & .75                                                               & 2.9  & 11      \\ \bottomrule
  \end{tabular}

  \caption*{The functional form defines the conditional probability of the machine's activation given the blickets (not non-blickets) that are on the machine, and it is a function of these blickets. In our experiment, the functional form can be interpreted as a rule that needs at least a threshold number of blickets to activate the blicket machine. At this threshold number, the activation occurs with some probability, but above the threshold, the activation always occurs. For example, with the noisy conjunctive form, the blicket machine activates with a .75 probability given a threshold of 2 blickets, but it always activates given 3 or more blickets. Each form has a corresponding sigmoid parameterization of bias and gain values.}
\end{table}

We presented participants with several active blicket tasks to investigate whether and how they would learn and transfer overhypotheses across these tasks.
The earlier training tasks were designed so that participants could easily learn overhypotheses. For example, the first training task had only 3 blocks, allowing participants to intervene on all $2^3=8$ possible combinations within the constraints of the task (the 45s time limit in Experiment 1, or the 12 intervention limit in Experiment 2).
The final transfer task was then designed to measure the transfer of overhypotheses learned from training.
The transfer task had more blocks (6 or 9) and thus was more combinatorially complex ($2^6 = 64$ or $2^9 = 512$ possible combinations), and it was no longer possible to intervene on all combinations of blocks within the task constraints. This complexity increased the importance of relying on previously learned overhypotheses to select just a few informative interventions. 

Throughout our two experiments, we performed between-subjects manipulations of the blicket machine's functional form in the training and transfer tasks, where all six functional forms we considered are listed in Table~\ref{tab:fforms}. We also manipulated the training length (one or two tasks). Our dependent measures were participants' interventions and causal judgments in the transfer task.
These measures would not only indicate whether participants' interventions and judgments were informative of the transfer task's causal structure and functional form, but also whether these were informative under overhypotheses about the functional form transferred from past training tasks. 

\subsection{Experiment 1}

\subsubsection{Participants}
\totalN{} participants were recruited using Amazon Mechanical Turk (HIT Approval Rate $\geq 99\%$, Number of HITs Approved $\geq 1000$, Age $\geq 18$) for the 8 between-subjects conditions in Fig.~\ref{fig:exp1_conditions}. From left to right in this figure, the number of participants in each condition is \dddN{}, \ddN{}, \ccdN{}, \cdN{}, \cccN{}, \ccN{}, \ddcN{}, and \dcN{}. The corresponding numbers of \textit{filtered} participants are: \fdddN{}, \fddN{}, \fccdN{}, \fcdN{}, \fcccN{}, \fccN{},  \fddcN{} and \fdcN{}. Participants were paid \$1.5 for completing the study (\completionTime{} on average, excluding the instructions) and received a bonus of up to \$1.05 for their questionnaire performance, resulting in a mean total compensation of \$\totalComp{}.

\subsubsection{Procedure}

Experiment 1 manipulated the functional form of the transfer task (disjunctive or conjunctive), whether this form was matched with their past training tasks (same or different), and training length (short or long, i.e., one or two training tasks before the final transfer task), creating 8 between-subjects conditions (Fig.~\ref{fig:exp1_conditions}). 

Within each task, participants saw a web interface with blocks (colored squares labeled with alphabetical letters) and a blicket machine. Examples of this interface are shown and described in Fig.~\ref{fig:tasks}. Participants were asked to identify which of the blocks were blickets with the help of the blicket machine, and they were told the blocks' colors, letters and positions did not reveal blickets. (Unknown to participants, we performed counterbalancing by randomizing the block color and whether a block was a blicket. Each block was labeled with a unique letter that was assigned in alphabetical order.) Participants could choose any number of interventions within a time limit of 45 seconds, where each intervention involved putting any combination of blocks on the machine. The machine would then respond by activating or doing nothing (according to a disjunctive or conjunctive functional form, which was unknown to participants). Participants could view their full history of interventions and machine responses. 

Participants encountered a first training task with three blocks. If they were in a \textit{long} training condition, they would encounter a second training task with six blocks followed by a final transfer task with nine blocks. Otherwise, if they were in a \textit{short} training condition, they would directly move on to the transfer task without seeing the second training task. The number of blocks in each task is also listed in Table~\ref{tab:exp1_nums}, along with how many of these blocks were blickets (unknown to participants). Even as the number of blocks (and along with it, the number of possible interventions) increased across tasks, the time limit remained at 45 seconds. 

\begin{table}[t]
  \centering
  \caption{Experiment 1: Number of blocks and blickets} 
  \label{tab:exp1_nums} 
  \begin{tabular}{lll} 
    \toprule
    Task       & Num. Total Blocks & Num. Blicket Blocks    \\
    \midrule
    Training 1 & 3                 & 1 (Disj.) or 2 (Conj.) \\
    Training 2 & 6                 & 3                      \\
    Transfer   & 9                 & 4                      \\      
    \bottomrule
  \end{tabular} 
  \caption*{In each task, the number of blickets is contained within the total number of blocks. In the first training task, the ``Disj.'' (Disjunctive) variant has one blicket while the ``Conj.'' (Conjunctive) variant has two.}
\end{table}

Each training and transfer task was followed by a questionnaire with two types of binary causal judgments, one about identifying each block as a blicket or non-blicket (``Which blocks do you think are blickets?''), and another about predicting whether the blicket machine would activate in the presence of different combinations of blocks (``Will the blicket machine activate (light up with a green color)?''; see Supplementary Methods for more details).

Between tasks, participants only received feedback and associated bonus compensation for the correctness of their activation prediction judgments, not their blicket identification judgments. We used this feedback/compensation structure to limit what was revealed about the ground truth causal relationship, since only getting feedback about whether or not a combination of blocks activated the machine would not reveal much about which of those blocks were blickets. Instead, participants would need to rely on their own interventions to identify blickets. Furthermore, the compensation would incentivize participants to make more accurate judgments, which meant they also needed to make more informative interventions.

\subsection{Experiment 2}
\subsubsection{Participants}
\twoTotalN{} participants were recruited using Amazon Mechanical Turk (HIT Approval Rate $\geq 99\%$, Number of HITs Approved $\geq 1000$, Age $\geq 18$) for the 6 between-subjects conditions in Fig.~\ref{fig:exp2_conditions}. From left to right in this figure, the number of participants in each condition is \twoDcN{}, \twoNdcN{}, \twoCcN{}, \twoNccN{}, \twoCccN{}, and \twoNcccN{}. Participants were paid \$1.28 for completing the study (\twoCompletionTime{} on average, excluding the instructions) and received a bonus of up to \$1.22 for their questionnaire performance, resulting in a mean total compensation of \$\twoTotalComp.

\subsubsection{Procedure}
Experiment 2 manipulated the functional form of the training task using all 6 forms in Table~\ref{tab:fforms}, creating 6 between-subjects conditions
(Fig.~\ref{fig:exp2_conditions}). The transfer task's functional form was fixed to the conjunctive form. 

The rest of the procedure was the same as Experiment 1 except for some adjustments to the training length, transfer task complexity, task constraints, and questionnaires. These differences are described below. 

While Experiment 1 used either one or two training tasks, Experiment 2 fixed the training length to a single training task. The transfer task's complexity was also reduced from 9 blocks to 6. This reduction addressed how a conjunctive transfer task with 9 blocks was likely too difficult to reveal whether participants' judgments were improving with matched overhypotheses (see Supplementary Results). See Table~\ref{tab:exp2_nums} for the number of blocks in each task, as well as how many of those are blickets. 

\begin{table}[t]
  \caption{Experiment 2: Number of blocks and blickets}
  \label{tab:exp2_nums}
  \centering
  \begin{tabular}{@{}lllll@{}}
    \toprule
    Task & Num. total blocks & Num. blickets & Functional form \\ \midrule
    \multirow{6}{*}{Training} & \multirow{6}{*}{3} & 1 & Disjunctive \\
      &  & 1 & Noisy Disjunctive   \\
      &  & 2 & Conjunctive         \\
      &  & 2 & Noisy Conjunctive   \\
      &  & 3 & 3-Conjunctive       \\
      &  & 3 & Noisy 3-Conjunctive \\
    \midrule
    Transfer & 6 & 3 & Conjunctive\\ \bottomrule
  \end{tabular}
  \caption*{In each task, the number of blickets is contained within the total number of blocks, and when these two numbers are the same, then all blocks are blickets. The training task always has three total blocks, but the number of blickets varies with the functional form.}
\end{table}

The constraint in each task was changed from a time limit (with a variable number of interventions that depended on the participant) to a fixed number of interventions: 12 in the training tasks, and 20 in the transfer task. The 12 training interventions allowed participants to learn both deterministic and noisy forms: while deterministic forms could be learned by intervening on all 8 combinations of 3 blocks, noisy forms could only be fully learned by repeating interventions, which was possible with the 4 remaining interventions. Our noisy forms had a .75 probability of activating given the threshold number of blickets (see Table~\ref{tab:fforms}), so on average, the participant could learn from one failed activation if their 4 repeated interventions included the threshold number of blickets. In the transfer task, the 20 intervention limit set a reasonable difficulty level relative to the 64 possible combinations of 6 blocks, while still requiring appropriate overhypotheses and informative interventions for fully learning the transfer task. Both the 12 (training) and 20 (transfer) intervention limit have been verified as sufficient under our hierarchical Bayesian model (see preregistration: \url{https://osf.io/vk9yd/?view_only=dd45fd1aafd9499d8173c64cae2deedc}). 

Following each task, the questionnaire had the same content as Experiment 1---blickets and the machine's activation---but differed in its exact format. The exact format is explained in our preregistration (\url{https://osf.io/vk9yd/?view_only=dd45fd1aafd9499d8173c64cae2deedc}). We do not go into details here because our analysis of Experiment 2 focuses on model evaluation rather than the questionnaire. However, the questionnaire was still important for evaluating participants' answers and awarding them a corresponding amount of bonus compensation. Participants did not receive any feedback about their answers until after the experiment. As in Experiment 1, this feedback/compensation structure required participants to learn from only their own interventions and incentivized them to choose more informative interventions. 

\subsection{Hierarchical Bayesian model and ablation models} \label{sec:HBM}
Our hierarchical Bayesian model represents causal beliefs at multiple levels of abstraction, including both lower-level beliefs about the causal structure and higher-level overhypotheses about the functional form. It infers the most likely causal structures and functional forms given the effects of different ensembles of blocks being placed on a blicket machine in the active blicket task. Each event is a pair $(q,o)$ of the intervention $q$ and the outcome $o$: the intervention is the set of blocks placed on the blicket machine, and the outcome is the binary response of the blicket machine (1 for activation or 0 for no activation). Given an event $(q,o)$, the full joint Bayesian update for a particular structure $s \in S$ and particular form $f \in F$ is:
\begin{align}
  \label{eq:joint}
  P(s,f|q,o) \propto P(q,o|s,f)P(s,f)
\end{align}

Each causal structure $s$ is represented by enumerating the set of blickets under this structure (e.g., \{A, B\} represents the causal structure where blocks A and B are blickets and any other blocks are non-blickets). The space of all causal structures $S$ in an active blicket task is the power set of all blocks in that task, and we used a uniform prior over $S$ for the start of each task. The space of functional forms $F$ is described in the next section.

\subsubsection{Commitment 1: People represent a rich space of overhypotheses}\label{sec:assump-1}

Our model's space of overhypotheses about functional forms follows from our experiment's cover story, which considers blickets as a general class of exchangeable objects that can have a generative effect (i.e., blicket machine activation). This means the functional form of the blicket machine's activation should only consider individual blickets important to the extent that they contribute to the overall \textit{number} of blickets that are on the machine: the functional form reduces to a function of the \textit{number} of blickets. Furthermore, because blickets are generative causes, the form should output a conditional probability value that monotonically increases with the number of blickets. These properties can be satisfied by any family of functions that maps the domain of zero and positive integers to the range $[0,1]$.

Therefore, following \citeauthor{lucasLearningFormCausal2010} (\citeyear{lucasLearningFormCausal2010}), our model considers the \textit{sigmoid} family of functional forms, evaluated at zero and positive integer inputs. This family is not only consistent with the exchangeability and generative properties of blickets, but it is also simple and able to express a rich space of forms with only two parameters, bias and gain. This space includes all the forms used in our experiment (see Table~\ref{tab:fforms} for the bias and gain values of our experiment's forms). It also includes gradations between these forms, where finer variations in blicket thresholds and noise levels can be thought of as variations of the bias and gain parameters, respectively.

A particular form in the sigmoid family is fully described by a pair of bias $b$ and gain $g$ values. It outputs the conditional probability of the machine's activation given the number of blickets $n$ on the machine:
\begin{align}
  \text{sigmoid}(n) = \frac{1}{1 + e^{-(g * (n - b))}}
\end{align}

Our model's space of overhypotheses about functional forms covers these combinations of $b$ and $g$:
\begin{align}
  b & \in \{0.15i | i \in \mathbb{Z} \land 0 \leq i \leq 19\} \\
  g & \in \{2j | j \in \mathbb{Z} \land 0 \leq j \leq 19\}
\end{align}

We chose gamma priors over bias and gain values that favor the kinds of functional forms people typically consider: disjunctive and reliable/nearly-deterministic forms \citep{lucasLearningFormCausal2010,luBayesianGenericPriors2008,mayrhoferSufficiencyNecessityAssumptions2016,schulzGodDoesNot2006}. We chose a varied grid of 24 priors that have these properties, as shown in Table~\ref{tab:prior_grid}, rather than only a single prior. We initialized and ran our model for each of these priors to produce predictions of people's interventions under each prior. We then marginalized over these priors in our model comparisons (see Section~\ref{sec:crossval_marg}).

\begin{table}[t]
  \caption{Bias and gain priors}
  \label{tab:prior_grid}
  \centering
  \begin{tabular}{@{}llllll@{}}
    \toprule
    \multicolumn{3}{c}{Bias prior} & \multicolumn{3}{c}{Gain prior} \\ \midrule
    Shape     & Scale    & Mode    & Shape     & Scale    & Mode    \\ \midrule
    4         & 0.1      & 0.3     & 101       & 0.1      & 10      \\
    4         & 0.1      & 0.3     & 11        & 1        & 10      \\
    4         & 0.1      & 0.3     & 201       & 0.1      & 20      \\
    4         & 0.1      & 0.3     & 21        & 1        & 20      \\
    2.2       & 0.25     & 0.3     & 101       & 0.1      & 10      \\
    2.2       & 0.25     & 0.3     & 11        & 1        & 10      \\
    2.2       & 0.25     & 0.3     & 201       & 0.1      & 20      \\
    2.2       & 0.25     & 0.3     & 21        & 1        & 20      \\
    6         & 0.1      & 0.5     & 101       & 0.1      & 10      \\
    6         & 0.1      & 0.5     & 11        & 1        & 10      \\
    6         & 0.1      & 0.5     & 201       & 0.1      & 20      \\
    6         & 0.1      & 0.5     & 21        & 1        & 20      \\
    3         & 0.25     & 0.5     & 101       & 0.1      & 10      \\
    3         & 0.25     & 0.5     & 11        & 1        & 10      \\
    3         & 0.25     & 0.5     & 201       & 0.1      & 20      \\
    3         & 0.25     & 0.5     & 21        & 1        & 20      \\
    9         & 0.1      & 0.8     & 101       & 0.1      & 10      \\
    9         & 0.1      & 0.8     & 11        & 1        & 10      \\
    9         & 0.1      & 0.8     & 201       & 0.1      & 20      \\
    9         & 0.1      & 0.8     & 21        & 1        & 20      \\
    4.2       & 0.25     & 0.8     & 101       & 0.1      & 10      \\
    4.2       & 0.25     & 0.8     & 11        & 1        & 10      \\
    4.2       & 0.25     & 0.8     & 201       & 0.1      & 20      \\
    4.2       & 0.25     & 0.8     & 21        & 1        & 20      \\ \bottomrule
  \end{tabular}
  \caption*{Grid of gamma priors for bias and gain. Each gamma prior is parameterized by the shape and scale, and is summarized by its mode.}
\end{table}

\subsubsection{Ablation model 1: Fixed-Form}
The Fixed-Form ablation model removes the commitment that people represent a rich space of overhypotheses. This ablation is implemented by replacing our hierarchical Bayesian model's space of sigmoid forms with a single deterministic disjunctive form.

\subsubsection{Commitment 2: People transfer their learned overhypotheses}
After our model's joint distribution over forms and structures is conditioned on events in one task, the posterior marginal distribution of functional forms is extracted and reused as the prior over functional forms for a new task. Our model then multiplies this transferred prior with a uniform distribution over the causal structures in the new task, creating a joint distribution for learning in the new task. Thus, our model predicts that people transfer their learned overhypotheses about functional forms from one task to the next.

The posterior marginal probability for a form $f$ is:
\begin{align}
  P(f|q,o) = \sum_{s \in S} P(s,f|q,o)
\end{align}
which is calculated from the joint posterior $P(s,f|q,o)$ that also includes causal structures $S$.

Transferring the marginal distribution of functional forms is possible because the likelihood of the joint inference (Equation~\ref{eq:joint}) is calculated hierarchically: 
The likelihood is proportional to the conditional probability (of the effect) defined by the functional form (higher-level overhypothesis), which depends on the causal structure (lower-level beliefs) as an input. We abuse the functional form notation $f$ to show this dependence in the likelihood:
\begin{align}
  P(q,o|s,f) = P(o|q,s,f)P(q|s,f) \propto
  \begin{cases}
    f(|q \cap s|)     & \text{if } o=1  \\
    (1-f(|q \cap s|)) & \text{if } o=0
  \end{cases}                      
\end{align}
where the form $f$ is a sigmoid function that returns the probability of activating the blicket machine ($o = 1$), given the number of blickets (set cardinality) in an intervention $q$ (a set of blocks on the blicket machine) according to the causal structure $s$ (a set of blocks that are blickets under this structure).

Because the functional form does not depend on any specific causal structure (i.e., the specific identities of non-blickets and blickets in any task), the same functional form can be used to compute likelihoods across distinct tasks with different non-blickets and blickets. The functional form only requires the abstract \textit{number} of blickets, which exists in all tasks.
Thus, the same space of functional forms, along with its marginal distribution, can be transferred for learning across tasks.

\subsubsection{Ablation model 2: No-Transfer}
The No-Transfer ablation model removes the commitment that people transfer their learned overhypotheses about the functional form. This ablation is implemented by discarding the part of our hierarchical Bayesian model that reuses the marginal posterior over functional forms from a previous task as the prior in a new task. Instead, it reinitializes the prior over functional forms so that it is the same at the start of every task. We considered a grid of priors, which are described in Section~\ref{sec:assump-1}.

\subsubsection{Commitment 3: People choose interventions that are informative for learning overhypotheses}
Like previous computational accounts of active causal learning \citep[e.g.,][]{steyversInferringCausalNetworks2003,bramleyConservativeForgetfulScholars2015,coenenStrategiesInterveneCausal2015}, our model considers an intervention's expected information gain ($\text{H}_{\text{gain}}$) with respect to causal structures $\mathbb{E}[\text{H}_{\text{gain}}(S|q,o)]$. Unlike previous accounts, our model additionally considers expected information gain with respect to overhypotheses about the functional form $\mathbb{E}[\text{H}_{\text{gain}}(F|q,o)]$. Our model prefers interventions that maximize expected information gain on both structures and forms, predicting that people choose interventions that are not only informative for learning the causal structure but also for learning overhypotheses about the functional form.

The expected information gain of an intervention $q$ for a random variable $X$ (functional forms $F$ or causal structures $S$) is:
\begin{align}
  \begin{split}
 & \mathbb{E}[\text{H}_{\text{gain}}(X|q,o)] = \\
 & \sum_{o \in \{0,1\}} \left[-\sum_{x \in X} P(x)\text{log}P(x) + \sum_{x \in X}P(x|q,o)\text{log}P(x|q,o) \right] P(o|q)
  \end{split}
\end{align}
where the outer expectation is calculated with the probability of each outcome $o$ for this intervention $q$. With this formulation, we note that preferring higher expected information gain is tantamount to preferring lower expected conditional entropy.

Rather than calculating expected information gain on the joint distribution over causal structures and functional forms, our model uses a linear combination of
their respective marginal expected information gains, weighted by parameter $w \in [0,1]$:
\begin{align}
  \label{eq:linear-combo}
  w\mathbb{E}[\text{H}_{\text{gain}}(F|q,o)] + (1-w)\mathbb{E}[\text{H}_{\text{gain}}(S|q,o)]
\end{align}
This equation is only equivalent to joint information gain when form and structure are independent conditional on interventions $q$ and outcomes $o$, which, here, they are not. However, we use this decomposition to approximate joint information gain because it serves the important purpose of allowing us to capture the possibility that people's interventions preferentially maximize one kind of information over another. When $w=0$, our model collapses to the special case of only trying to learn about structure, as in previous models of active causal learning, while $w=1$ implies that only the functional form matters, as we might expect if people are only interested in learning overhypotheses for future use.

\subsubsection{Ablation model 3: Structure-Only-EIG}
The Structure-Only-EIG ablation model removes the commitment that people choose interventions that are informative for learning overhypotheses. This ablation is implemented by setting $w=0$ (Equation~\ref{eq:linear-combo}) in our hierarchical Bayesian model, predicting that interventions are only informative (i.e., maximizing expected information gain) for learning causal structures and not overhypotheses about functional forms.

\subsubsection{Random baseline model}
Our final comparison model is a random baseline that samples interventions uniformly. For example, in a task with 6 blocks, the random baseline model would sample an intervention from $2^6=64$ possible combinations of blocks with a probability of $\frac{1}{64}$ for each combination.

\subsection{Model comparison}
We compared our hierarchical Bayesian model, the ablation models and the random baseline model according to how well they predicted participant interventions in Experiment 2's transfer task. We did not perform this comparison on Experiment 1's transfer task because of the associated computational cost. In Experiment 2's transfer task, each of the \twoTotalN{} participants made 20 interventions, each of which was a choice between the 64 possible combinations of 6 blocks. Experiment 2's 6-block transfer task had a reasonable cost for computing model predictions for every intervention (given a prior, a model ran for up to 22 total hours on one 2018 Apple MacBook Pro CPU (2.3 GHz Quad-Core Intel Core i5)). In contrast, Experiment 1's 9-block transfer task was much more costly: predicting a single intervention by a single participant (for a total of \totalN{} participants) would involve computations on 512 possible combinations of 9 blocks. Thus, we only computed our model predictions and evaluations on Experiment 2's transfer task.

\paragraph{Predictive likelihood}
To evaluate how well each model predicted participant interventions, we calculated the predictive likelihoods of participant interventions under that model. This predictive likelihood is the probability that the model would have chosen that intervention. Thus, if a model assigns higher predictive likelihoods to participant interventions, it is better at predicting the participant's intervention choice.

Our hierarchical Bayesian model and the ablation models assign a predictive likelihood to a participant intervention $q_p$ by: (1) conditioning the model's joint distribution over forms and structures on that participant's history of interventions (up to and not including $q_p$) and outcomes in the current task, (2) calculating $q_p$'s combined expected information gain (on both causal structures and functional forms; see Equation~\ref{eq:linear-combo}), and (3) applying the softmax function $\sigma$. Let the combined expected information gain be abbreviated to cEIG, then the predictive likelihood of $q_p$ is:

\begin{align}
  \sigma(\text{cEIG}(q_p)) = \frac{e^{\frac{1}{t}\text{cEIG}(q_p)}}{\sum_{i} e^{\frac{1}{t}\text{cEIG}(q_i)}}
\end{align}
where $t$ is the temperature parameter of the softmax function. $i$ enumerates all the \textit{possible} choices for that intervention, as opposed to (but does include) the actual choice $q_p$ made by the participant. In Experiment 2's transfer task, a participant made 20 interventions and at each intervention, they had 64 \textit{possible} choices, corresponding to all the possible combinations of blocks.

The softmax temperature $t$ controls how sensitive the model is to predicting (i.e., assigning higher predictive likelihoods to) intervention choices that have higher cEIG: lower temperatures increase the model's preference for intervention choices that maximize cEIG, while higher temperatures decrease the model's preference for any intervention choice, making the model tend toward assigning a uniform predictive likelihood to all possible intervention choices. 

The random baseline model samples interventions uniformly, so it assigns a fixed predictive likelihood of $\frac{1}{64}$ to every participant intervention in Experiment 2's transfer task. 64 is the number of possible choices for any intervention, corresponding to all the possible combinations of blocks.

\subsubsection{Cross-validation and prior marginalization}\label{sec:crossval_marg}
In order to calculate the predictive likelihood under our model and the ablation models (and not the random baseline model), two parameters need to be fitted: the softmax temperature $t$ and the weight $w$ in the combined expected information gain (see Equation~\ref{eq:linear-combo}). It is not clear what kinds of temperatures and weights would be appropriate for predicting people's behavior, so we fit them to participant interventions using the training folds in cross-validation. We considered all temperature-weight combinations from the values below:
\begin{align}
  t & \in \{0.001, 0.01, 0.1, 1, 10, 100\} \\
  w & \in \{0.1, 0.2, 0.3, 0.4, 0.5, 0.6, 0.7, 0.8, 0.9, 1.0\}
\end{align}
The parameterization $w=0$ (not listed above) corresponds to removing the second commitment that people choose interventions that are informative for learning overhypotheses. Thus, it is the \textit{only} weight value considered in the second ablation model and it is not considered for any other model.

For each model, we performed two kinds of cross-validation evaluations: an \textit{averaged} evaluation over all participants ($N=\twoTotalN$) and interventions (20 per participant) in Experiment 2's transfer task, and an \textit{individual differences} evaluation that only considered one participant's interventions at a time.

In the averaged evaluation, participants were randomly split into four balanced folds. Holding out one fold at a time, we fitted a model's parameters by maximizing the mean predictive likelihood in the remaining folds, where this mean is first calculated for each functional form prior (Table~\ref{tab:prior_grid}) and then marginalized over all priors using a uniform distribution (this marginalization is not applicable to the first ablation model, which only has a single prior with a single disjunctive form). We then used these fitted parameters to evaluate the model's mean predictive likelihood in the hold-out fold with the same process for marginalizing over priors. The mean of all four hold-out evaluations was used for comparing models, where higher values meant a model was a better predictor of average participant intervention strategies.

In the individual differences evaluation, the cross-validation process was the same as the averaged one except it was performed within each individual participant. For each participant, their 20 interventions were split into four balanced folds and the mean of the hold-out evaluations was used for selecting the best model for that participant. If a model was the best predictor for a higher number of participants, then that model was a better predictor of individual intervention strategies.

\bibliography{refs}%

\newpage
\begin{appendices}
  
\section{Supplementary Results}\label{sec:suppl-results}
\subsection{Experiment 1}
\subsubsection{Causal judgments}

Aside from blicket identification judgments, we also considered another type of causal judgment in the transfer task: 7 predictions about whether or not a combination of blocks would activate the blicket machine (see Section~\ref{sec:suppl-methods} for more details). Like for the blicket identification judgments, we expected the activation prediction accuracy to be predicted by the match between the transfer and training functional forms, training length, and their interaction, considering the transfer task's functional form as a covariate. We used these variables to fit a logistic regression model to predict the per-participant accuracy percentage in the transfer task activation prediction questions (binomial with 7 trials).
The main effect of the match of functional form was not significant in the full data ($z=1.24, p=.215$), but was significant for the filtered participants who were more engaged with the transfer task ($z_f=2.53, p_f=.012$).
The transfer task's functional form had a significant main effect ($z=3.67, p<.001$; filtered: $z_f=4.23, p_f<.001$), which was consistent with past results suggesting that people find disjunctive forms easier to learn \citep{lucasLearningFormCausal2010}.
Surprisingly, the length of training and its interaction with the match between training and transfer forms were not significant predictors.

\begin{figure}[t]
  \centering
  \includegraphics[width=\textwidth]{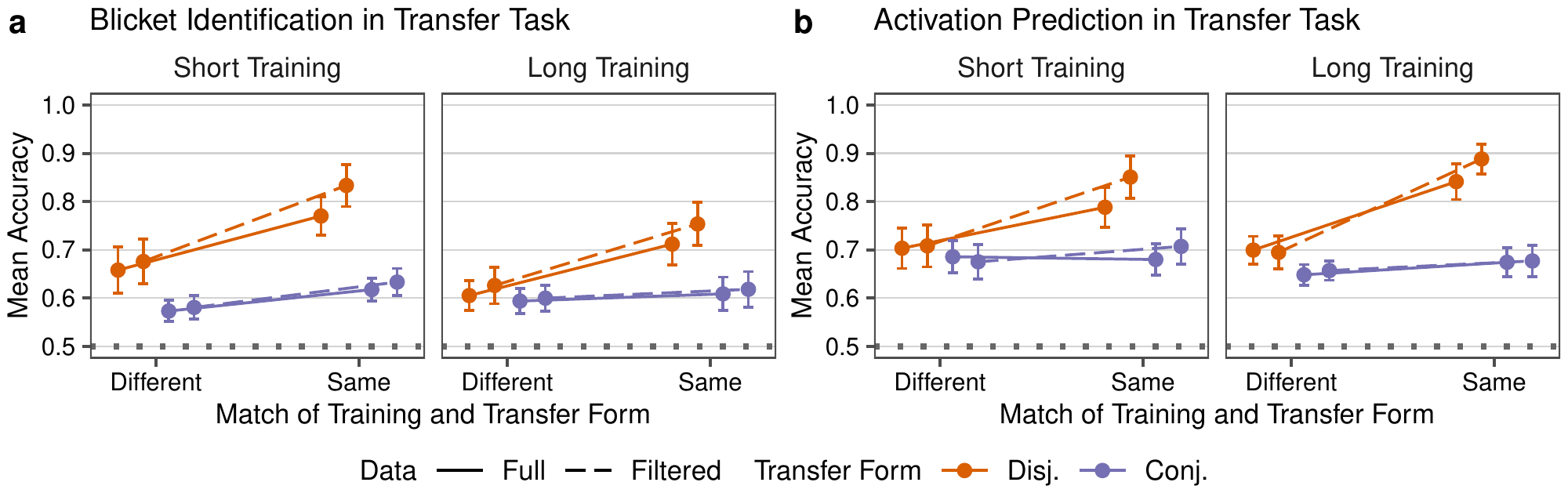}
  \caption{Experiment 1: Questionnaire performance in the transfer task, grouped by the transfer functional form (``Disj.'' for Disjunctive, or ``Conj.'' for Conjunctive), its match with the training form (Same or Different), and training length (Long or Short). Chance ($.5$) accuracy is shown with a dotted gray line. Error bars in either direction denote the magnitude of the standard error. Mean participant accuracies for \textbf{a} blicket identification and \textbf{b} activation prediction are calculated separately for the full and filtered data.}
  \label{fig:exp1_quiz}
\end{figure}

We also used Welch t-tests (two-tailed) to investigate the specific effects of match between pairs of conditions (visualized in Fig.~\ref{fig:exp1_quiz}b), expecting causal judgment accuracies to improve from mismatched to matched conditions.
In the disjunctive transfer conditions, the comparisons were consistent with our expectations: Mean activation prediction accuracy improved significantly from mismatched to matched conditions with long training, $t(50.00)=-3.04, p=.004$ (filtered: $t_f(42.37)=-4.18, p_f<.001$). The short training improvement was not significant in the full data ($t(52.70)=-1.45, p=.154$), but was significant in the filtered data ($t_f(43.89)=-2.31, p_f=.025$).
In the conjunctive transfer conditions, however, the difference between matched (conjunctive training) and mismatched (disjunctive training) accuracies was non-significant. This weaker match effect might have accounted for the non-significant interaction effect between match and training length in our logistic regression model.
We suspected this weaker effect was due to the conjunctive transfer task being too difficult to learn, regardless of training match and length. This suspicion was supported by the blicket identification results in our next experiment, where we lowered the difficulty of the conjunctive transfer task and found a significant improvement from mismatched to matched conditions (see Section~\ref{sec:addr-exp1}).

\subsubsection{First intervention}
\begin{figure}[t]
  \centering
  \includegraphics[width=\textwidth]{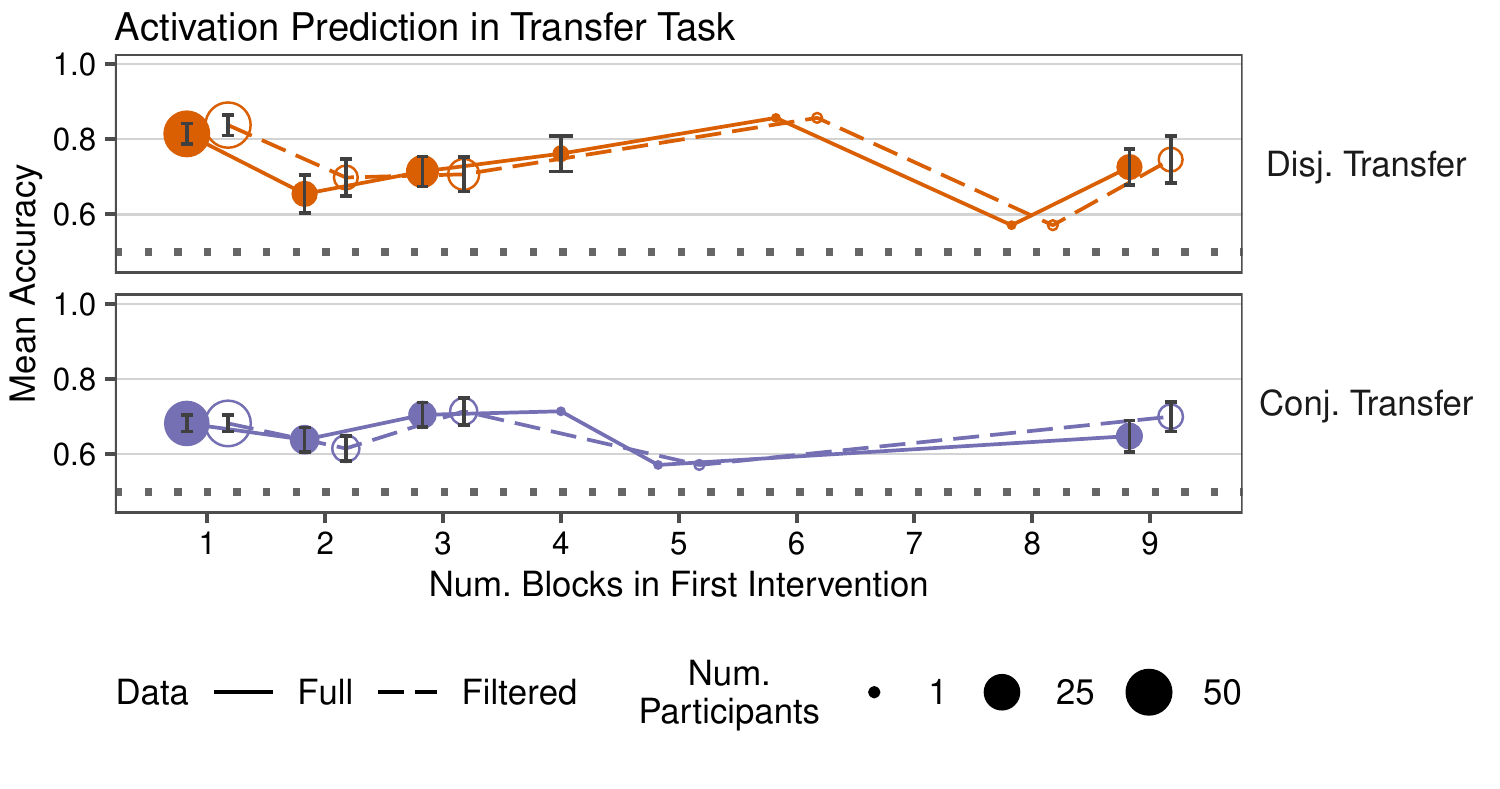}
  \caption{Experiment 1: Mean participant accuracies for activation prediction questions in the transfer task. This is grouped by the number of blocks in the first intervention and the transfer form. The mean is calculated separately for the full (solid lines) and filtered (dashed lines) data. Error bars in either direction denote the magnitude of the standard error but are omitted for points with a single participant, where the standard error is ill-defined. Chance (.5) accuracy is shown with a dotted gray line.}
  \label{fig:exp1_predAcc_numBlocks}
\end{figure}

To understand when the first intervention would be efficient for learning in the transfer task, we fitted a (binomial) logistic regression model to predict activation prediction accuracy (7 trials). The predictors included the number of blocks in the first intervention, the functional form of the transfer task, and their interaction.
There was a significant main effect of the transfer form ($z=3.38, p<.001$; filtered: $z_f=4.29, p_f<.001$), but no other significant effects (all $p \geq .078$). Our figures suggest that even though participants were able to identify a larger subset of blickets with efficient interventions (see Fig~\ref{fig:exp1_blicketAcc_numBlocks} in the main text), this partial knowledge was not sufficient to make more accurate activation predictions (Fig.~\ref{fig:exp1_predAcc_numBlocks}), which had a larger coverage over blickets and their combinations with other blocks.

\subsection{Addressing Experiment 1's Non-Significant Results}\label{sec:addr-exp1}

In Experiment 1, the conjunctive transfer task conditions had a non-significant difference in blicket judgments between matched (conjunctive) and mismatched (disjunctive) training for both long (2 tasks) and short (1 task) training lengths. We had expected a strong effect of matched vs. mismatched training, especially in the longer training conditions that gave additional opportunities to learn about the matched or mismatched form. Instead, we found a non-significant trend. One possible explanation would be that participants were not learning conjunctive overhypotheses through training, and thus, participants with matched conjunctive training were performing no better than those with mismatched disjunctive training. However, this explanation seems unlikely and it is possible participants were fatigued or frustrated due to the difficulty of Experiment 1's conjunctive transfer task, which involved finding 4 blickets among 9 blocks (which can be intervened on in $2^9=512$ ways) within 45 seconds. For example, a simple and reasonable strategy under a conjunctive form would be to intervene on only the 36 possible \textit{pairs} of blocks, but even this was not possible under the short time limit. In contrast, the same time limit allowed participants to successfully find blickets in the easier disjunctive variant of the transfer task by testing all 9 singleton blocks. Indeed, performance was lower in the conjunctive transfer task conditions than in the disjunctive transfer task conditions ($p < .001$ for both blicket classification and activation prediction questions in the full and filtered data). 

\begin{figure}
  \centering
  \includegraphics[width=0.45\textwidth]{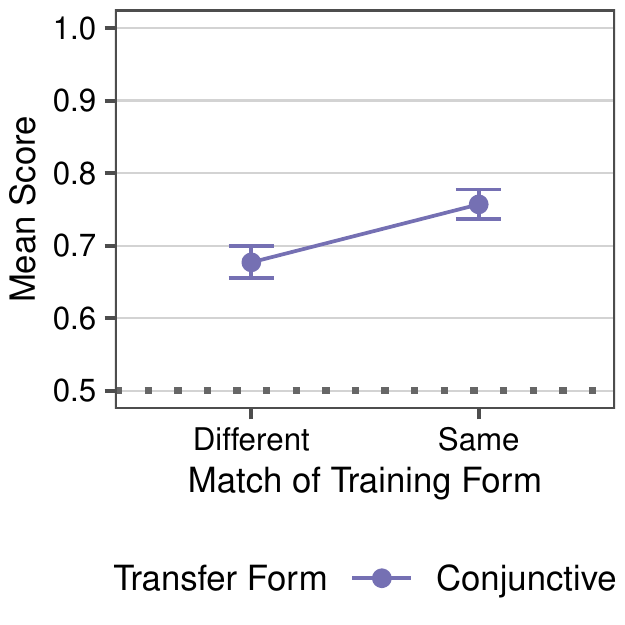}
  \caption{Experiment 2: Mean blicket rating score in the conjunctive transfer task. The plotted mean scores are grouped by whether the training functional form was the same (deterministic conjunctive condition) or different (deterministic disjunctive condition). A participant's blicket rating (0-10) is scored as $1 - \frac{|\text{participant rating} - \text{true rating}|}{10}$, where the true rating is 10 for blicket blocks and 0 for non-blicket blocks. These scores are averaged over participants and each of their 6 ratings (one for each block) in the transfer task. The chance level score ($.5$) is shown with a dotted gray line and error bars in either direction denote the magnitude of the standard error. This plot shows participants' scores in the conjunctive transfer task improved from mismatched to matched training.}
  \label{fig:exp2_conjtransfer_blicketscore}
\end{figure}

To address the difficulty of Experiment 1's conjunctive transfer task, we designed an easier version for Experiment 2, asking participants to find 3 blickets among 6 blocks (which can be intervened on in $2^6=64$ ways) with a fixed intervention number of 20. In this easier conjunctive transfer task, we now found a significant improvement in mean blicket judgments from mismatched to matched training (two-tailed Welch t-test: $t(85.856)=-2.17, p=.033$; comparison is visualized in Fig.~\ref{fig:exp2_conjtransfer_blicketscore}), where the mismatched and matched training correspond to the deterministic disjunctive and deterministic conjunctive training conditions in Experiment 2. Thus, Experiment 2's results address Experiment 1's non-significant differences in blicket judgments.

Experiment 2's conjunctive transfer task had a few more differences compared with Experiment 1's. Experiment 2 asked participants to rate blickets on a 0-10 scale instead of asking them to classify them on a binary scale (blicket or not). The blicket rating scale follows \citeauthor{lucasLearningFormCausal2010}'s (\citeyear{lucasLearningFormCausal2010}) measure of disjunctive versus conjunctive training effects in a passive learning setting and allows us to measure these effects more precisely in our active learning setting as well. Experiment 2 also did not vary the training length like Experiment 1, but instead used only a short (1 task) training in all its conditions. However, since we already see a significant effect in Experiment 2's short training, we expect that this effect would remain or be larger with longer training. 

\section{Supplementary Methods}\label{sec:suppl-methods}
\subsection{Experiment 1}

In addition to blicket identification questions, the questionnaire after each task also included binary predictions of whether the blicket machine would activate in the presence of different combinations of blocks (``Will the blicket machine activate (light up with a green color)?''). There were seven different predictions about seven different combinations of blocks, including combinations with zero, one and two blickets along with other non-blicket blocks (where the number and identities of blickets and non-blickets were unknown to participants), as well as one combination with all blocks in the task.
For example, consider a transfer task with the nine blocks \{J, K*, L*, M, N, O, P, Q*, R*\} (where blickets are marked with an asterisk for the sake of this example). The seven combinations (subject to randomization of the exact blickets and non-blickets) could then be \{N, O, Q*\}, \{J, M, R*\} (one blicket); \{P, Q*, R*\}, \{K*, L*, O\} (two blickets); \{J, N, O\}, \{M, N, O, P\} (zero blickets); and finally one combination containing all nine blocks.

\end{appendices}

\end{document}